\documentclass{article}
\usepackage{iclr2027_conference,times}
\usepackage[T1]{fontenc}

\usepackage{bm}
\usepackage[hyphens]{url}
\usepackage{graphicx}
\usepackage{amsmath,amssymb}
\usepackage{booktabs,multirow}
\usepackage{subcaption}
\usepackage{xcolor}

\usepackage{xspace}

\newcommand{\method}{OIC-GS}
\newcommand{\eg}{e.g.,\xspace}
\newcommand{\ie}{i.e.,\xspace}

\newcommand{\rpg}{RDPS}
\newcommand{\rcc}{RCC}

\title{Rate-Distortion Adaptive Primitive Selection\\ for Omnidirectional Gaussian Splatting}

\author{
Yulong~Cheng$^{1}$\quad
Youneng~Bao$^{2}$\quad
Junfeng~Zhou$^{1}$\quad
Mu~Li$^{1}$\thanks{Corresponding author.}\quad
Jie~Wen$^{1}$\\
$^{1}$Harbin Institute of Technology, Shenzhen, China\\
$^{2}$Shenzhen University, Shenzhen, China\\
\texttt{dark.stven@gmail.com},\ \texttt{baoyn@szu.edu.cn},\ \texttt{limu403@163.com}
}

\iclrfinalcopy

\begin{document}

\maketitle
\lhead{Preprint}

\begin{abstract}
Learned image codecs (LICs) achieve high reconstruction quality, but their decoding speed is often insufficient for immersive virtual reality (VR). Gaussian splatting (GS) codecs render much faster, yet still lag in reconstruction quality and typically decide primitive allocation
without considering the coding cost of each primitive.
We introduce \method{}, an omnidirectional GS codec with a new hierarchical HEALPix primitive grid representation. Gaussian primitives are anchored at predefined spherical locations, eliminating explicit coordinate coding. Finer levels refine their coarser ancestors, naturally supporting coarse-to-fine reconstruction and layered transmission. The predefined grid also enables efficient viewport decoding by selecting only view-relevant primitives.
We further introduce a lightweight entropy model for quantized primitives and optimize the codec under a spherical rate-distortion objective. Primitives with insufficient rate-distortion benefit are automatically removed when their quantized opacity becomes zero, allowing \method{} to adapt both primitive density and level of detail without a fixed primitive budget.
A single bitstream supports full-sphere, viewport-dependent, and progressive decoding. The first viewport reaches final quality after decoding only $52\%$ of the bitstream, and is then rendered at $1{,}270$\,FPS. On a 100-image omnidirectional benchmark, \method{} outperforms all evaluated GS codecs, reducing WS-PSNR BD-rate by $49.6\%$ over GaussianImage++ and $68.6\%$ over SGI, which uses a learned entropy model.

\end{abstract}

\section{Introduction}
\label{sec:intro}

Immersive virtual reality (VR) requires high-resolution omnidirectional content with low decoding latency. Although learned image codecs (LICs) achieve strong reconstruction quality, they typically require entropy decoding of the full latent representation followed by a decoding network, making real-time decoding challenging. Gaussian splatting (GS) \citep{kerbl2023_3dgs} offers a more efficient alternative by reconstructing images through lightweight rasterization of explicit primitives.

However, existing GS codecs for omnidirectional images are typically built on the equirectangular projection (ERP), whose highly non-uniform sampling density causes severe oversampling near the poles. As a result, optimization tends to allocate excessive primitives to these oversampled regions, leading to inefficient bitrate allocation and degraded rate-distortion (R-D) performance. Existing approaches also commonly adopt a two-stage optimization strategy \citep{li2025gaussianimagepp,zhu2025lig,sgi2026}: the primitive locations or numbers are first determined based on reconstruction quality, and the resulting primitive set is then fixed while its attributes are optimized. This separation prevents the coding cost and distortion reduction of each primitive from being jointly considered, making R-D-optimal primitive selection difficult.

VR viewing introduces another important requirement. Unlike conventional images, an omnidirectional image is viewed only through a local viewport at any given time. Existing learned and GS codecs nevertheless reconstruct the entire panorama before projecting the requested viewport, introducing unnecessary transmission and decoding latency. A more efficient solution is to transmit and decode only the primitives required by the current viewport, while progressively delivering the remaining content when needed.

To address these issues, we propose \method{} (Fig.~\ref{fig:overview}), a content-adaptive hierarchical GS codec that is fitted to each panorama following the overfitted coding paradigm \citep{coin2021,strumpler2022inr,ladune2023coolchic,kim2024c3} and optimized end-to-end under a rate-distortion (R-D) objective. Primitives are organized on hierarchical HEALPix grids \citep{Gorski2005HEALPix}, where each level progressively refines its coarser ancestors. Since primitive positions are predefined by the spherical grid, only their existence needs to be coded, eliminating the overhead of explicitly coding coordinates. Moreover, HEALPix operates directly on the sphere with approximately uniform sampling, avoiding the severe sampling imbalance introduced by ERP. A lightweight entropy model estimates the rate of primitive attributes
using Gaussian mixtures conditioned on previously decoded hierarchy levels
as a prior and on a spherical checkerboard context within each level. Primitive existence is jointly optimized through quantized opacity, such that a primitive is selected only when its distortion reduction justifies its estimated coding cost. The resulting hierarchical bitstream supports two complementary forms of progressive decoding: coarse-to-fine refinement across hierarchy levels and viewport-to-full-sphere decoding, where only the primitives required by the current viewport are transmitted and decoded first. With a tile-aligned storage format that increases file size by only $1.2\%$ on average, a single bitstream supports full-sphere, viewport-dependent, and progressive streaming decoding (App.~\ref{app:viewport}).

In summary, our main contributions are threefold:
\begin{itemize}
    \item \textbf{Rate-distortion adaptive primitive selection.} We couple
    the quantized opacity of each primitive with the bitrate estimated by
    the entropy model and optimize the resulting rate jointly with
    distortion. This encourages primitives with limited distortion
    reduction to gradually reduce their opacity to zero, enabling automatic
    primitive selection within a unified R-D optimization framework.
    \item \textbf{One bitstream, three decoding modes, and real-time viewport rendering.} \method{} supports full-sphere, viewport-dependent, and progressive decoding, all bit-exact within the viewport; the first viewport needs only $52\%$ of the file.
    \item \textbf{A new hierarchical HEALPix GS representation for single omnidirectional image compression.} Built directly on the sphere, the proposed representation avoids ERP sampling imbalance and enables hierarchical coarse-to-fine coding. Experiments show that \method{} outperforms all tested GS image codecs across all bitrates, reducing WS-PSNR BD-rate by $49.6\%$ compared with GaussianImage++.
\end{itemize}

\section{Related Work}
\label{sec:related}
In this section, we briefly review omnidirectional image compression and compression methods based on Gaussian splatting.

\paragraph{Omnidirectional image compression.}
Panoramas are conventionally projected onto a plane (\eg ERP, cubemap
or hybrid projections) and compressed with planar image or video codecs
\citep{wallace1992jpeg,skodras2001jpeg2000,sullivan2012hevc,VVC}, where
padding or content-adaptive projections reduce the projection
distortion \citep{he2018content,xu2020state}. Learned image codecs
\citep{balle2018variational,minnen2018joint,Cheng2020,elic} have also
been extended to omnidirectional images. An end-to-end optimized
$360^\circ$ codec \citep{Li2022EndToEnd} allocates bits adaptively to
regions at different ERP latitudes, and pseudocylindrical
representations \citep{li2026pseudocylindrical} sample the sphere
nearly uniformly, so that standard convolutions can be applied with
pseudocylindrical padding. OSLO \citep{Bidgoli2022OSLO} instead learns
directly on the sphere, building on HEALPix-based spherical networks
such as DeepSphere \citep{Perraudin2019DeepSphere}. More recently, a
viewport-based codec \citep{viewport360nic2026} replaces the projection
with a fixed set of extracted viewports, which are coded one after
another under a cross-viewport context model. These
methods decode a static image in full, or a fixed chain of viewports,
before the requested viewport is shown, so their decoding latency
becomes significant at high resolutions. The addressable precincts of
JPEG2000 \citep{jpip} provide region-wise access, and $360^\circ$ video
is delivered with viewport-adaptive streaming
\citep{corbillon2017viewport}, but both operate on the projection
plane, where a viewport near the poles becomes a full-width band. In
contrast, \method{} renders a viewport directly from the primitives at
display resolution, without a full-image network pass, and its tiles
are equal-area HEALPix cells, so a polar viewport is not expanded into
a full-width band. Once decoded, the viewport is rendered in real time,
which supports interactive viewing.

\paragraph{Gaussian splatting compression.}
Adapted from 3D reconstruction \citep{kerbl2023_3dgs}, 2DGS
\citep{zhang2024gaussianimage} represents an image with colored
Gaussians rendered by accumulated blending, and a growing number of
methods improve this representation
\citep{imagegs2025,instantgi2025,zhu2025lig,contour2dgs2025}, including
layered progressive coding on the plane \citep{pgsvc2026}. In these works, attribute compression is driven by reconstruction
quality rather than by a probability model: they learn their quantizers but use
fixed entropy coding or bit counting
\citep{zhang2024gaussianimage,li2025gaussianimagepp,zhang2024_2dgsic,sga2dgs2025},
so the number of primitives cannot be driven by a rate
gradient--GaussianImage++ \citep{li2025gaussianimagepp} densifies
primitives according to reconstruction error and LIG
\citep{zhu2025lig} allocates them by level of detail, neither comparing
a primitive against the bits it would cost. 

To our knowledge, SGI \citep{sgi2026} is the only GS image codec with a
learned entropy model. Its probability model, however, is conditioned
on primitive locations through a hash grid that must itself be
transmitted, and the number of primitives is fixed a priori rather than
determined by the rate term. Both limitations also appear in 3D Gaussian splatting compression,
where conditional entropy models over anchor attributes
\citep{hac2024,contextgs2024} and learnable masks
\citep{compact3dgs2024,wang2024rdogaussian} have been studied: HAC
\citep{hac2024} queries a hash grid at the anchor position, and its
mask loss counts the retained Gaussians, so pruning any Gaussian saves
the same amount regardless of its actual code length. We instead
condition the probability model on the rendering state that the decoder
has already reconstructed, which costs no side information, and let the
code length assigned to each primitive drive its existence. Following
the overfitted coding paradigm
\citep{coin2021,ladune2023coolchic,kim2024c3}, \method{} fits one model
per image and entropy-codes its quantized parameters into the
bitstream.

\section{Method}
\label{sec:method}
In this section, we first present the hierarchical spherical Gaussian
representation of \method{}, including its quantization, the
rate-distortion adaptive primitive selection (\rpg{}) and the rendering
process (Sec.~\ref{sec:field}). We then describe the entropy and rate model
based on the rendering-conditioned context (\rcc{}, Sec.~\ref{sec:entropy}),
and finally the joint rate-distortion (R-D) optimization
(Sec.~\ref{sec:alloc}).

\subsection{Hierarchical Spherical Gaussian Representation}
\label{sec:field}

\begin{figure}[t]
\centering
\includegraphics[width=\linewidth]{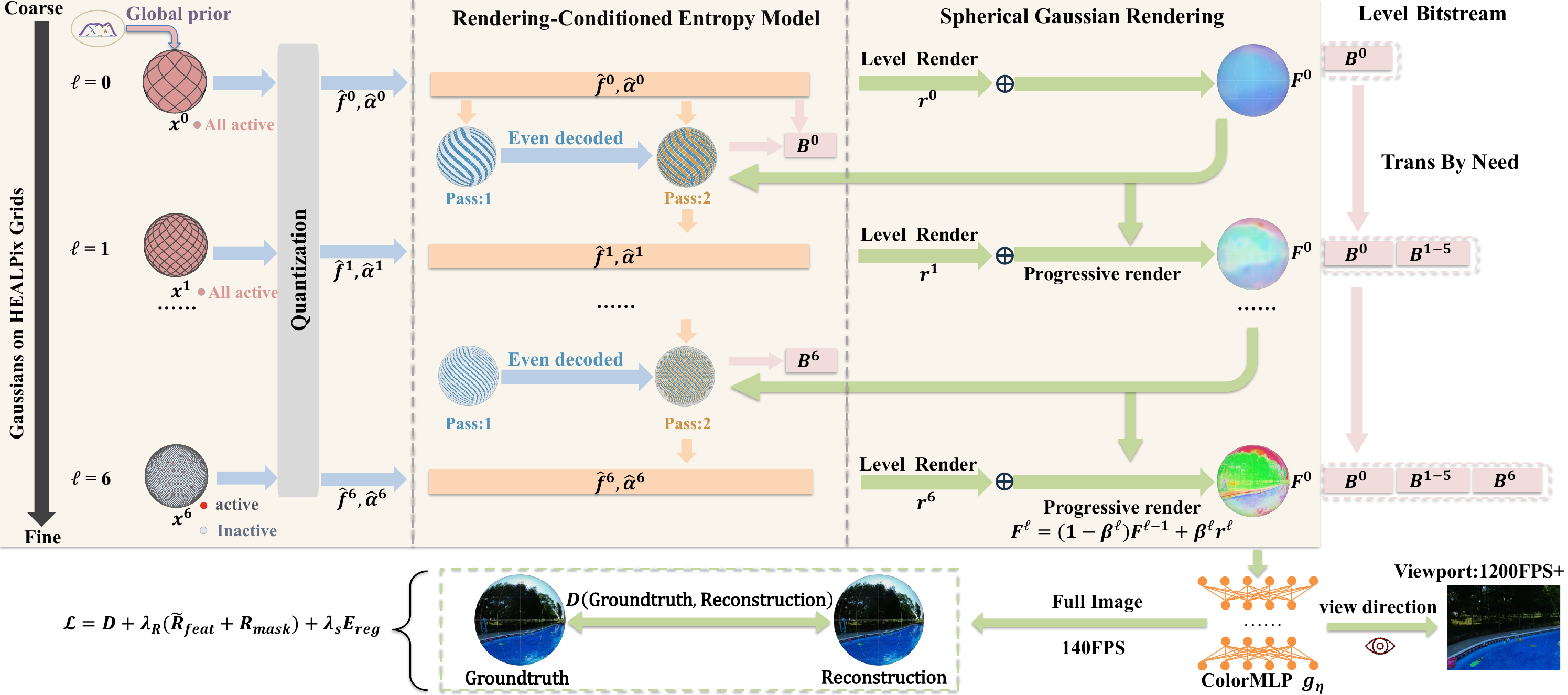}
\caption{Overview of \method{}. Primitives are anchored on hierarchical
HEALPix levels and carry a quantized feature and opacity, coded level by
level. The levels are rendered from coarse to fine; a primitive whose
quantized opacity is zero becomes inactive, and its cell falls back to the
coarser levels. The rendering state $(F^{(\ell)},r^{(\ell)})$ is
reconstructed identically at the encoder and the decoder.}
\label{fig:overview}
\end{figure}

\noindent\textbf{Representation.}
\method{} represents an omnidirectional image as a hierarchy of primitives
$X=\{\bm{x}^{(0)},\dots,\bm{x}^{(L-1)}\}$ on the sphere, where $L=7$ by
default and each level is an equal-area HEALPix grid that holds one
primitive per cell (Fig.~\ref{fig:overview}). Denoting the cells of level
$\ell$ by $\Omega^{(\ell)}$, with
$|\Omega^{(\ell)}|=12\cdot(4\cdot2^{\ell})^2$, we have
$\bm{x}^{(\ell)}=\{x^{(\ell)}_i\mid i\in\Omega^{(\ell)}\}$. With NESTED
indexing, each cell at level $\ell-1$ exactly covers four cells at level
$\ell$. Every primitive is anchored at its cell center and carries two
attributes, $x^{(\ell)}_i=(f^{(\ell)}_i,\alpha^{(\ell)}_i)$: the feature
$f^{(\ell)}_i$ is what it contributes to the reconstruction, and the
opacity $\alpha^{(\ell)}_i$ acts as an existence variable that decides
whether the primitive is kept. With this fixed-position design, a
primitive has no position or covariance of its own: its position is given
by the grid, and all primitives of a level share one isotropic kernel
width.

\noindent\textbf{Quantization.}
Both attributes are quantized with the uniform $b$-bit quantizer
\begin{equation}
\label{eq:quant}
Q_b(u)=\Delta_b\Bigl\lfloor\frac{u}{\Delta_b}\Bigr\rceil,\qquad
\Delta_b=\frac{1}{2^b-1},
\end{equation}
which gives the features $\hat f^{(\ell)}_i=Q_4(f^{(\ell)}_i)$ with the
quantization step $\Delta_f=\Delta_4$, and the opacities
$\hat\alpha^{(\ell)}_i=Q_2(\alpha^{(\ell)}_i)\in\{0,\tfrac13,\tfrac23,1\}$.
During training, the rounding is bypassed with the straight-through
estimator (STE) \citep{bengio2013ste} in the reconstruction and replaced by
additive uniform noise \citep{balle2017endtoend} in the rate estimate.

\noindent\textbf{Primitive selection and existence mask.}
The existence of a primitive is determined by its quantized opacity
through the mask
\begin{equation}
\label{eq:mask}
m^{(\ell)}_i=\mathbf{1}\bigl[\hat\alpha^{(\ell)}_i>\tau\bigr],\qquad \tau=0.2.
\end{equation}
Selection is a binary decision. However, deactivating a primitive within a single update removes its
contribution at once, while the neighboring primitives and the coarser
levels have not yet adapted to cover its region. Instead, we let a primitive leave the representation gradually.
Among the four opacity states, only $\hat\alpha^{(\ell)}_i=0$ falls below
$\tau$, \ie the two intermediate states $\{\tfrac13,\tfrac23\}$ act as a
buffer between being active and inactive, in which a primitive still
contributes with a reduced opacity, allowing the neighboring primitives
and the coarser levels to adapt gradually. Since the opacities are
optimized under the R-D objective of Sec.~\ref{sec:alloc}, the mask, and hence the set of active primitives, is determined by
rate-distortion optimization rather than by a preset primitive budget; we
refer to this mechanism as rate-distortion adaptive primitive selection
(\rpg{}). Only the active primitives are rendered with their own
attributes and written into the bitstream.

\noindent\textbf{Rendering.}
Given a direction $(\theta,\phi)$ with unit vector $v(\theta,\phi)$, the
levels are rendered from coarse to fine, so that the rendering state
$F^{(\ell-1)}(v)$ of the coarser levels is available when level $\ell$ is
rendered, with $F^{(-1)}\equiv0$. Since the rendering of a direction combines several neighboring primitives
of each level, an inactive primitive that contributed nothing would leave
the result determined only by the remaining active neighbors, which may
misrepresent the content of its region. We therefore let an inactive
primitive take its attributes from the coarser levels:
\begin{equation}
\label{eq:gate}
\bigl(\tilde\alpha^{(\ell)}_i,\,\tilde f^{(\ell)}_i\bigr)
=m^{(\ell)}_i\bigl(\hat\alpha^{(\ell)}_i,\,\hat f^{(\ell)}_i\bigr)
+\bigl(1-m^{(\ell)}_i\bigr)\bigl(\alpha_{\mathrm{fill}},\,\bar F^{(\ell-1)}_i\bigr),
\end{equation}
where $\bar F^{(\ell-1)}_i$, the coarse-level prediction, is the average of
$F^{(\ell-1)}$ over cell $i$, and $\alpha_{\mathrm{fill}}$ is a fixed value. At level $\ell$, only the
cell containing $v$ and its 8 HEALPix neighbors, denoted
$\mathcal{C}^{(\ell)}(v)$, are involved, and the feature of level $\ell$ at
$v$ is the normalized weighted average
\begin{equation}
\label{eq:splat}
r^{(\ell)}(v)=\frac{\sum_{i\in\mathcal{C}^{(\ell)}(v)} w_{v,i}\,\tilde f^{(\ell)}_i}
{\sum_{i\in\mathcal{C}^{(\ell)}(v)} w_{v,i}},
\end{equation}
where the weight $w_{v,i}$ is the product of $\tilde\alpha^{(\ell)}_i$ and a
spherical Gaussian kernel of the great-circle distance between $v$ and the
center of cell $i$ (App.~\ref{app:repro_model}). The opacity thus plays two
roles: its quantized value decides the existence of the primitive, and its
effective value weights the contribution of the primitive in
Eq.~\eqref{eq:splat}. The per-level results are accumulated from coarse to
fine,
\begin{equation}
\label{eq:blend}
F^{(\ell)}(v)=\bigl(1-\beta^{(\ell)}(v)\bigr)F^{(\ell-1)}(v)
+\beta^{(\ell)}(v)\,r^{(\ell)}(v),
\end{equation}
where the blending weight $\beta^{(\ell)}(v)\le0.5$ ensures that the final
image is never carried by the finest level alone. Finally, the ColorMLP
$g_\eta$ maps the final state of each direction to its RGB color,
\begin{equation}
\label{eq:color}
\hat{Y}(\theta,\phi)=g_\eta\bigl(F^{(L-1)}(v(\theta,\phi))\bigr),
\end{equation}
where $\hat Y$ denotes the reconstructed panorama. Since any direction can
be rendered in this way, the same procedure renders either the full sphere
or only the pixel directions of a viewport (App.~\ref{app:viewport}). As
all levels share one feature space and a single ColorMLP, and each level
only refines the coarser state, the state after any level can also be
decoded by $g_\eta$, which yields a coarse-to-fine preview from partially
received levels.

\subsection{Entropy and Rate Model}
\label{sec:entropy}
\begin{figure}[t]
\centering
\includegraphics[width=0.88\linewidth]{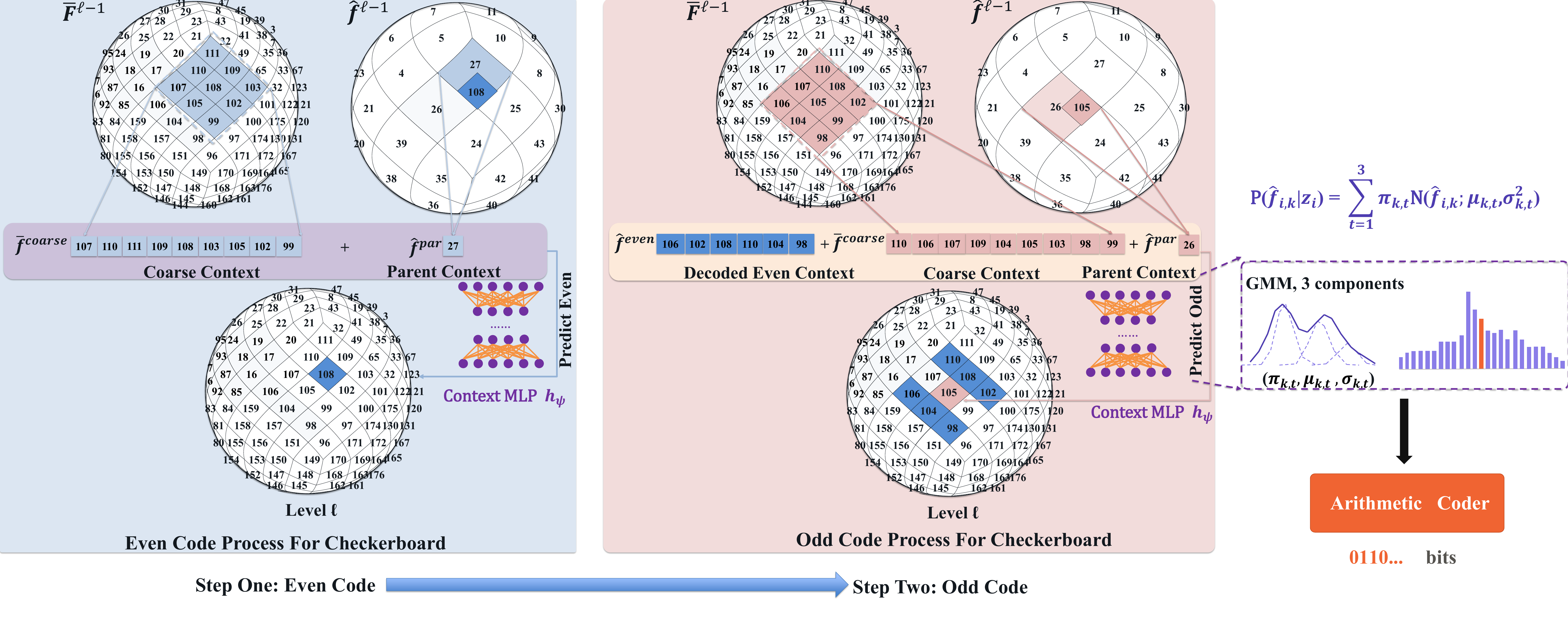}
\caption{The rendering-conditioned entropy model (\rcc{}). Even cells
(pass~1) are coded with the coarse and parent contexts; the decoded even
cells are then added to the context of the odd cells (pass~2). The
ContextMLP $h_\psi$ predicts a 3-component Gaussian mixture for the
arithmetic coder, and the decoder reconstructs the identical context by
construction. Cell indices are illustrative.}
\label{fig:cb}
\end{figure}

\noindent\textbf{Rendering-conditioned context.}
The features are coded level by level from coarse to fine. Before primitive
$x^{(\ell)}_i$ is decoded, all coarser levels have been decoded and
rendered, and this rendering state serves as its context, which we call
the rendering-conditioned context (\rcc{}, Fig.~\ref{fig:cb}). The context
consists of three parts,
\begin{equation}
\label{eq:ctx_main}
z^{(\ell)}_i=\bigl\{c_{\mathrm{coarse}}(\ell,i),\,
c_{\mathrm{par}}(\ell,i),\,
c_{\mathrm{even}}(\ell,i)\bigr\}.
\end{equation}
The coarse context $c_{\mathrm{coarse}}(\ell,i)$ consists of the
coarse-level predictions $\bar F^{(\ell-1)}$ of Eq.~\eqref{eq:gate} at cell
$i$ and its 8 neighbors, denoted $\mathcal{B}_9(i)$; they have the
resolution of level $\ell$ and lie in the same feature space as the coded
feature, and thus serve directly as a prediction reference. The parent
context $c_{\mathrm{par}}(\ell,i)$ is the decoded feature of the parent
cell, which covers the same region at the next coarser scale. The even context
$c_{\mathrm{even}}(\ell,i)$ follows the two-pass schedule of
\citet{he2021checkerboard}, adapted to HEALPix: the cells of each level are
split by the parity of their NESTED index (App.~\ref{app:rate}), the even
cells are decoded in the first pass, with $c_{\mathrm{even}}$ replaced by
the corresponding coarse values, and the odd cells in the second pass, with
$c_{\mathrm{even}}$ given by the decoded features of their even neighbors.
Since every part of the context is computed from previously decoded
symbols, the decoder reconstructs it identically without any side
information.

\noindent\textbf{Weighted entropy model.}
The $k$-th channel of the quantized feature $\hat f^{(\ell)}_i$ is modeled
by a discretized 3-component Gaussian mixture,
\begin{equation}
\label{eq:gmm}
P\bigl(\hat f^{(\ell)}_{i,k}\mid z^{(\ell)}_i\bigr)
=\int_{\hat f^{(\ell)}_{i,k}-\Delta_f/2}^{\hat f^{(\ell)}_{i,k}+\Delta_f/2}
\sum_{t=1}^{3}\pi_{k,t}\,\mathcal{N}\bigl(\xi;\mu_{k,t},\sigma^2_{k,t}\bigr)\,d\xi,
\end{equation}
where the weights $\pi_{k,t}$, means $\mu_{k,t}$ and variances
$\sigma^2_{k,t}$ are produced by the ContextMLP $h_\psi$ from
$z^{(\ell)}_i$; level 0, which has no coarser level, uses a global prior
GMM instead. Since we model the channels as conditionally independent given $z^{(\ell)}_i$, the
code length of a primitive and the feature rate of the representation are
\begin{equation}
\label{eq:rate_main}
R^{(\ell)}_i=-\frac{1}{N_{\mathrm{ERP}}}\sum_{k=1}^{C}\log_2
P\bigl(\hat f^{(\ell)}_{i,k}\mid z^{(\ell)}_i\bigr),\qquad
R_{\text{feat}}=\sum_{\ell=0}^{L-1}\sum_{i\in\Omega^{(\ell)}}
m^{(\ell)}_i R^{(\ell)}_i,
\end{equation}
where $N_{\mathrm{ERP}}$ is the number of ERP pixels, so that rates are
measured in bits per pixel (bpp), and the mask keeps only the active
primitives, whose features are written with an adaptive arithmetic coder
driven by these mixtures. In training, the binary mask is replaced by a
fraction of the quantized opacity, so that a primitive in the buffer of
Sec.~\ref{sec:field} is charged only part of its rate. The fraction is the
piecewise-linear function
\begin{equation}
\label{eq:pw}
\omega(\hat\alpha)=
\begin{cases}
\dfrac{\hat\alpha}{\tau}\,\omega_{\text{lo}}, & \hat\alpha\le\tau,\\[6pt]
\omega_{\text{hi}}+\dfrac{(\hat\alpha-\tau)(1-\omega_{\text{hi}})}{1-\tau}, & \hat\alpha>\tau,
\end{cases}
\end{equation}
which reduces to $\omega(\hat\alpha)=\hat\alpha$ with the default constants
$\omega_{\text{lo}}=\omega_{\text{hi}}=\tau$, and the charged feature rate
used in training is
\begin{equation}
\label{eq:rfeat_w}
\widetilde R_{\text{feat}}=\sum_{\ell=0}^{L-1}\sum_{i\in\Omega^{(\ell)}}
\omega\bigl(\hat\alpha^{(\ell)}_i\bigr)\,R^{(\ell)}_i.
\end{equation}

\noindent\textbf{Mask rate.}
In the proposed representation, the mask determines which cells of each
level hold an active primitive, and therefore has to be coded as well. We
model the mask entries of level $\ell$ as i.i.d.\ Bernoulli variables,
$m^{(\ell)}_i\sim\mathrm{Bernoulli}\bigl(\bar m^{(\ell)}\bigr)$, whose
parameter is the selection ratio
$\bar m^{(\ell)}=|\Omega^{(\ell)}|^{-1}\sum_i m^{(\ell)}_i$ of the level.
The mask rate is then
\begin{equation}
\label{eq:rmask_main}
R_{\text{mask}}=\frac{1}{N_{\mathrm{ERP}}}
\sum_{\ell=0}^{L-1}|\Omega^{(\ell)}|\,H\bigl(\bar m^{(\ell)}\bigr),\qquad
H(p)=-p\log_2p-(1-p)\log_2(1-p).
\end{equation}
The masks are written with a static binary arithmetic coder under the same
per-level probability, so that $R_{\text{mask}}$ matches the written mask
bits up to the header overhead. The nonzero opacities of the active
primitives take only three values and are coded with a static frequency
table. The bitstream stores, level by level from coarse to fine, the mask,
the nonzero opacities and the features of each level
(App.~\ref{app:repro_bits}).

\subsection{Rate-Distortion Optimization}
\label{sec:alloc}

All parameters of the image, \ie the continuous features and opacities of
the primitives, the ColorMLP, the ContextMLP, and the per-level kernel and
blending parameters, are jointly fitted by minimizing
\begin{equation}
\label{eq:rd}
\mathcal{L}=D+\lambda_R\bigl(\widetilde R_{\text{feat}}+R_{\text{mask}}\bigr)
+\lambda_s E_{\text{reg}},
\end{equation}
where $D$ is the mean squared error between the rendered colors and the
target colors, \ie the input panorama resampled to the equal-area target
grid, so that the sphere is uniformly weighted, $\lambda_R$ controls the
trade-off between quality and bitrate, and $E_{\text{reg}}$ is a light
regularizer on the per-level kernel widths and blending parameters
(App.~\ref{app:repro_model}). In this objective, the opacity of each
primitive is balanced between the distortion reduction it provides and its
own code length (Eq.~\eqref{eq:rfeat_w}), so that the number and placement
of the active primitives emerge from the optimization.

\section{Experiments}
\label{sec:exp}
\subsection{Experimental Setups}
\label{sec:settings}
Our main benchmark consists of 100 panoramas at $2048{\times}1024$ ERP
resolution, stored as PNG, from the Flickr-sourced $360^\circ$ dataset
of \citet{li2026pseudocylindrical}; for cross-dataset evaluation, we
use all 100 images of the SUN360 \citep{xiao2012sun360} test set
provided by \citet{deng2021lau}. We report WS-PSNR
\citep{sun2017wspsnr} under the JVET common test conditions
\citep{jvet}, and V-PSNR, V-SSIM \citep{wang2004ssim} and V-LPIPS
\citep{zhang2018unreasonable} over six $90^\circ{\times}90^\circ$
viewports; planar ERP metrics are not used, since the polar caps take
$33\%$ of the ERP pixels for $13\%$ of the sphere, and bitrates are
measured in bits per ERP pixel. Every GS baseline is fitted per image
with its public implementation
\citep{zhang2024gaussianimage,li2025gaussianimagepp,zhu2025lig,sgi2026,zhang2024_2dgsic}
(settings in App.~\ref{app:repro_eval}), and JPEG
\citep{wallace1992jpeg}, JPEG2000 \citep{skodras2001jpeg2000} and COIN
\citep{coin2021} serve as reference codecs.

\subsection{Rate-Distortion Performance}
\label{sec:rd}
\begin{figure}[htbp]
\centering
\includegraphics[width=0.96\linewidth]{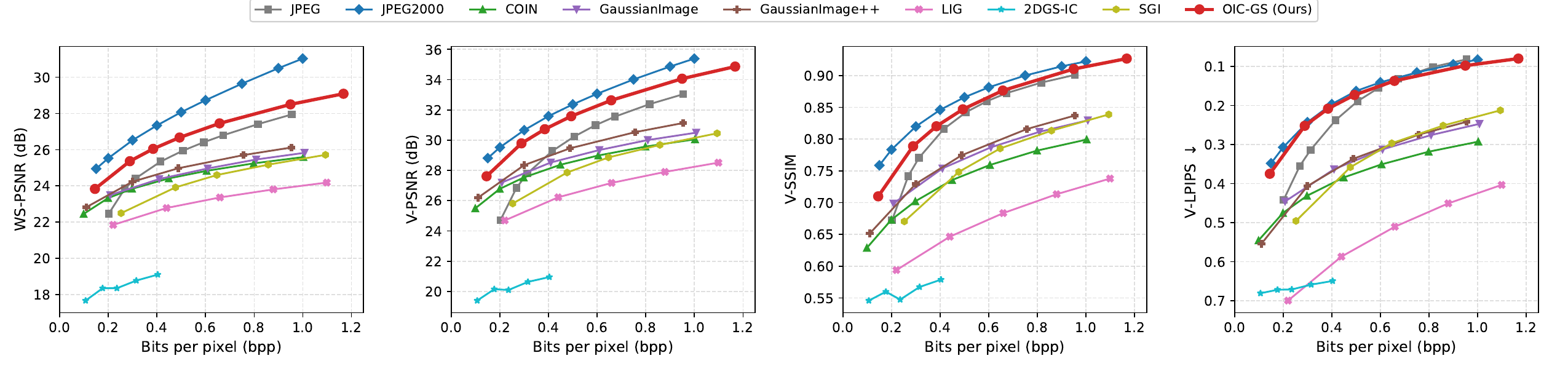}
\caption{Rate-distortion performance on the 100-image main benchmark.
\method{} outperforms every GS codec at every tested bitrate on all
four metrics.}
\label{fig:rd}
\end{figure}

\begin{figure}[htbp]
\centering
\includegraphics[width=0.96\linewidth]{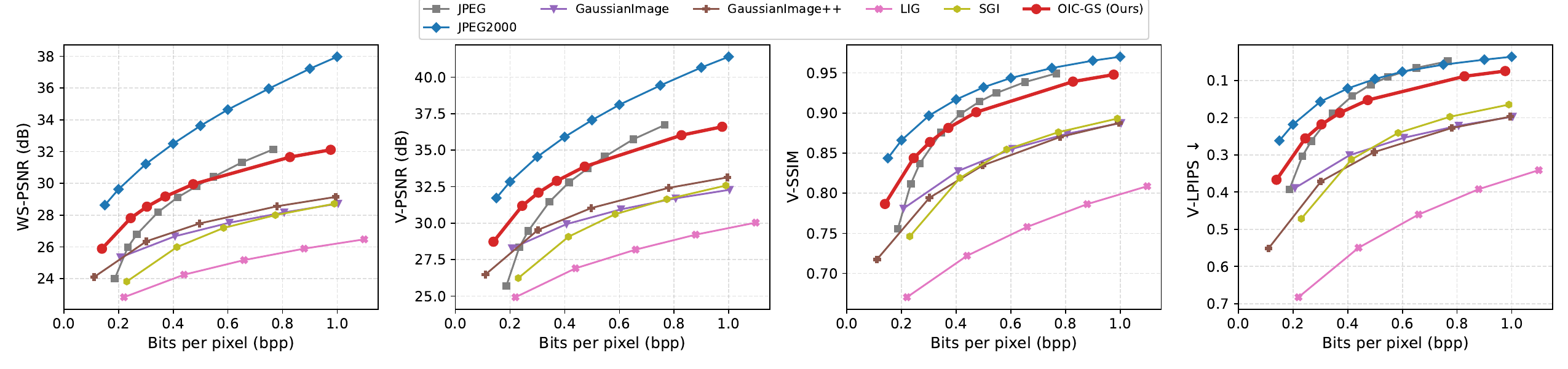}
\caption{Rate-distortion performance on the 100 SUN360 images.
\method{} again outperforms every GS codec at every tested bitrate on
all four metrics.}
\label{fig:crossdata}
\end{figure}

As shown in Figs.~\ref{fig:rd} and \ref{fig:crossdata} and
Table~\ref{tab:bdrate}, \method{} outperforms all GS codecs at all
tested bitrates on all four metrics, on both the main benchmark and
SUN360. Against SGI, the only existing GS codec with a learned entropy model, it
reduces the BD-rate by $68.6\%$, $67.1\%$, $57.7\%$ and $66.3\%$ on
WS-PSNR, V-PSNR, V-SSIM and V-LPIPS, respectively, and against
GaussianImage++ by $49.6\%$ on WS-PSNR
(Table~\ref{tab:pairwise}).

\begin{figure}[htbp]
\centering
\includegraphics[width=0.9\linewidth]{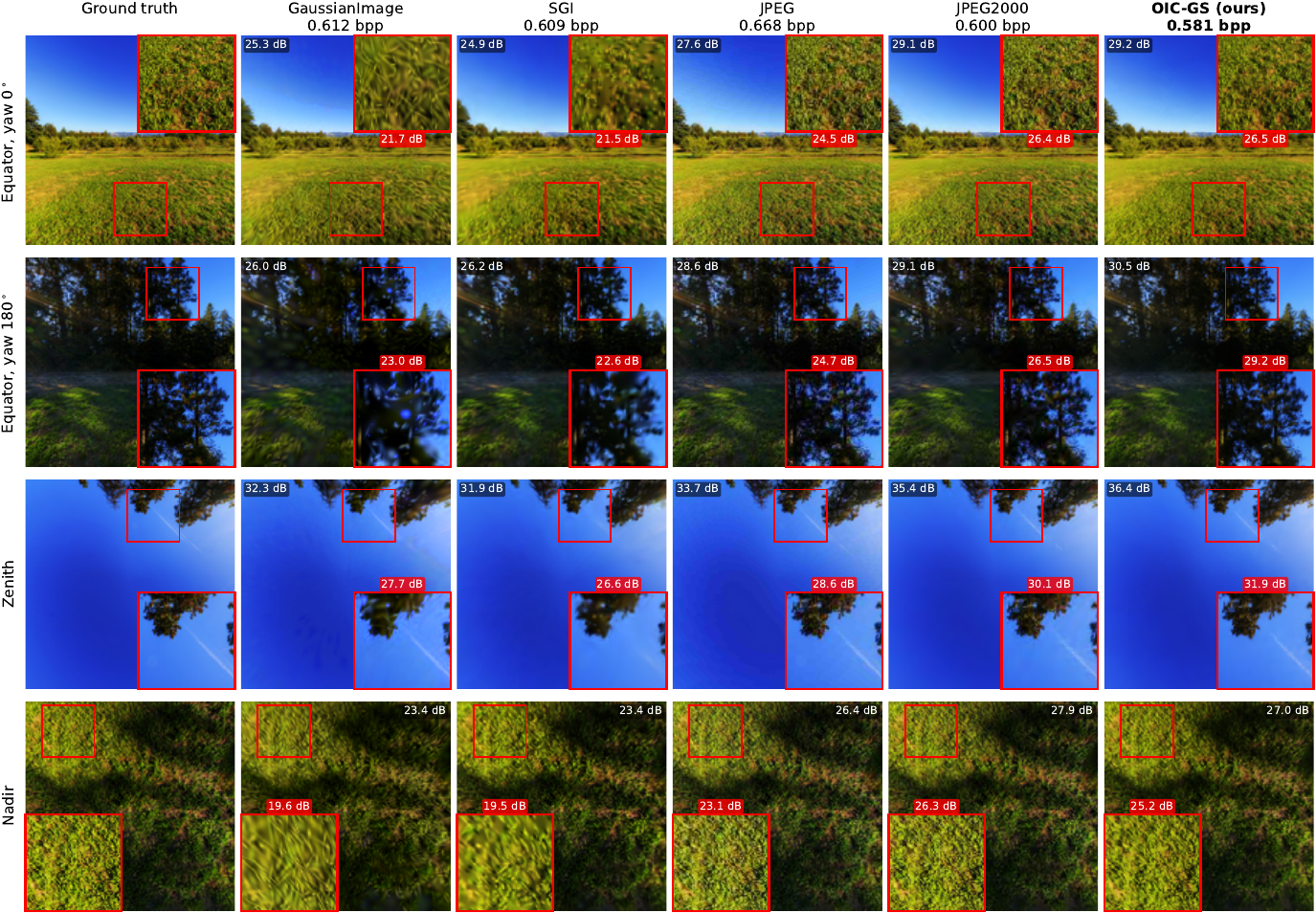}
\caption{Reconstructed viewports. Every baseline uses at least as many bits
as \method{}. The labels give the viewport PSNR and, for each $2\times$
zoomed-in patch, the PSNR of the textured $128^2$ window.
App.~\ref{app:qualitative_full} shows five scenes in all six directions.}
\label{fig:qualitative}
\end{figure}

\begin{table}[!t]
\centering
\small
\setlength{\tabcolsep}{4pt}
\begin{tabular}{lcccccccc}
\toprule
& \multicolumn{4}{c}{Main benchmark} & \multicolumn{4}{c}{SUN360} \\
\cmidrule(lr){2-5}\cmidrule(lr){6-9}
Method & WS-PSNR & V-PSNR & V-SSIM & V-LPIPS & WS-PSNR & V-PSNR & V-SSIM & V-LPIPS \\
\midrule
\multicolumn{9}{l}{Gaussian splatting image codecs} \\
SGI  & $+77.6$ & $+54.0$ & $+73.1$ & $+89.9$ & $+84.2$ & $+69.5$ & $+76.1$ & $+95.2$ \\
GaussianImage               & $+42.3$ & $+22.9$ & $+49.8$ & $+68.0$ & $+69.0$ & $+56.4$ & $+67.5$ & $+80.2$ \\
GaussianImage++             & $+5.5$ & $-9.5$ & $+31.3$ & $+66.5$ & $+26.8$ & $+16.9$ & $+68.1$ & $+103.6$ \\
LIG                         & $+165.9$ & $+95.4$ & $+254.2$ & $+369.7$ & $+214.0$ & $+159.3$ & $+308.7$ & $+396.2$ \\
\textbf{\method{} (ours)}   & $\mathbf{-29.6}$ & $\mathbf{-35.8}$ & $\mathbf{-17.2}$ & $\mathbf{-20.3}$ & $\mathbf{-13.3}$ & $\mathbf{-14.4}$ & $\mathbf{-4.1}$ & $+0.5$ \\
\midrule
\multicolumn{9}{l}{\emph{Non-GS reference codecs}} \\
COIN                        & $+15.4$ & $+2.9$ & $+69.5$ & $+102.1$ & $+28.6$ & $+18.3$ & $+94.1$ & $+125.0$ \\
JPEG2000                    & $-52.4$ & $-46.3$ & $-30.3$ & $-22.0$ & $-55.4$ & $-47.1$ & $-28.3$ & $-21.8$ \\
\bottomrule
\end{tabular}
\caption{BD-rate \citep{bjontegaard2001} against JPEG on the main benchmark and on the 100 SUN360 images; a negative value means fewer bits at equal quality.}
\label{tab:bdrate}
\end{table}

\subsection{Decoding Runtime and Qualitative Evaluation}
\label{sec:runtime}

\begin{figure}[t]
\centering
\includegraphics[width=0.88\linewidth]{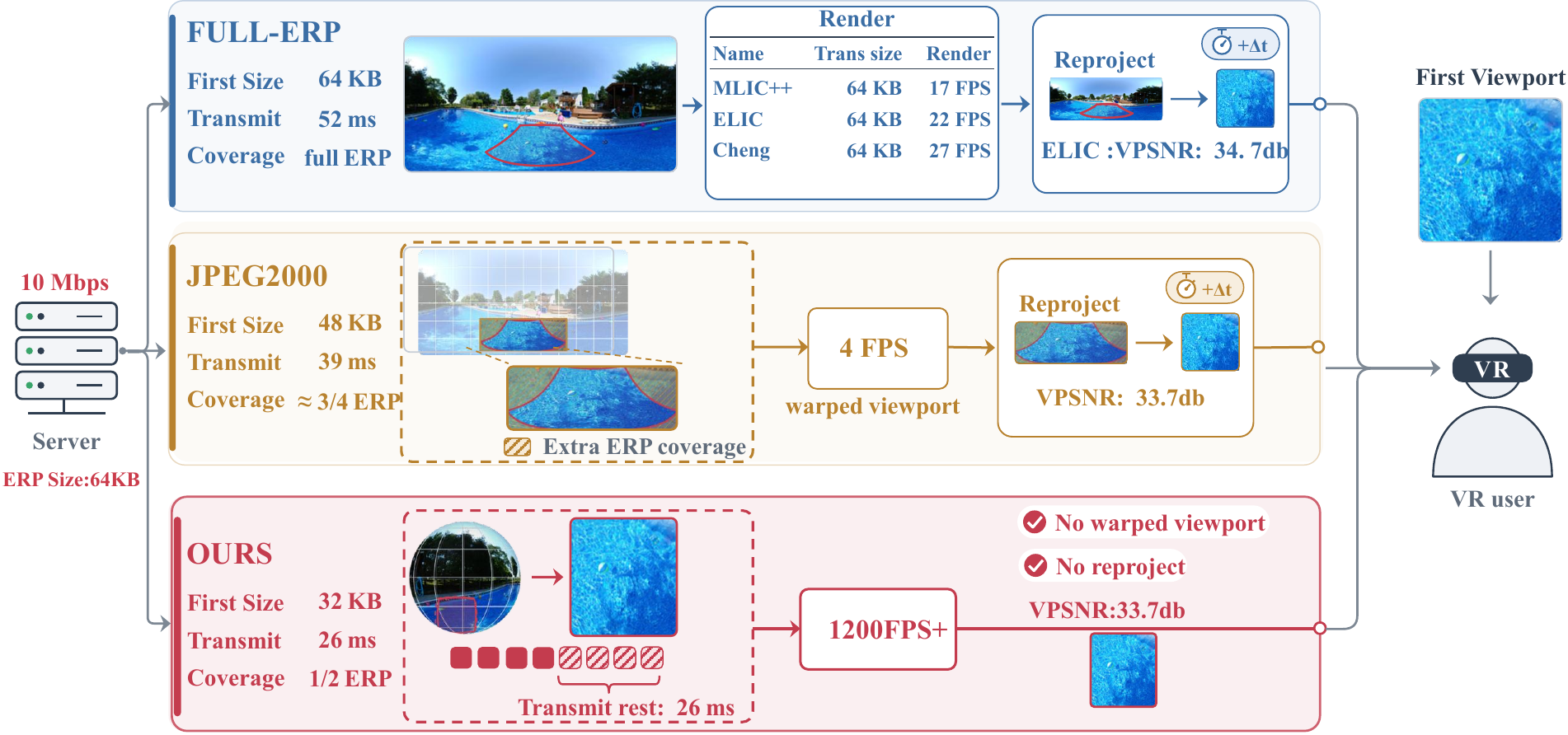}
\caption{Schematic comparison of first-viewport delivery over a
10\,Mbps link for full-ERP codecs, tiled JPEG2000 and \method{}.
Size: bytes transmitted before the first viewport is shown; Transmit:
transfer time of these bytes.}
\label{fig:three_view_compare}
\end{figure}

\paragraph{Decoding runtime.}
For panorama viewing, the latency before the first viewport matters more
than the speed of full reconstruction. \method{} displays a viewport
without the whole file or an ERP reprojection
(Fig.~\ref{fig:three_view_compare}): the first viewport reaches final
quality from $52\%$ of the file, bit-exact with full decoding, and is
rendered at $1{,}270$\,FPS (Table~\ref{tab:decode_fps}). Further details
are given in App.~\ref{app:viewport}.

\begin{table}[t]
\centering
\small
\setlength{\tabcolsep}{4pt}
\begin{tabular}{llrr}
\toprule
Method & Category & Entropy decoding (ms) & Reconstruction (FPS) \\
\midrule
JPEG \citep{wallace1992jpeg}                  & Traditional                  & \multicolumn{2}{r}{$97$} \\
JPEG2000 \citep{skodras2001jpeg2000}          & Traditional                  & \multicolumn{2}{r}{$3.8$} \\
\midrule
ELIC \citep{elic}                             & LIC, checkerboard context    & $466$ & $22.1$ \\
MLIC++ \citep{jiang2025mlicpp}                & LIC, multi-reference context & $621$ & $17.4$ \\
mbt2018 \citep{minnen2018joint}               & LIC, autoregressive          & $30{,}480$ & $67.1$ \\
cheng2020-attn \citep{Cheng2020}              & LIC, autoregressive          & $30{,}490$ & $27.4$ \\
\midrule
GaussianImage \citep{zhang2024gaussianimage}  & 2D GS                        & -- & $2{,}706$-$3{,}224$ \\
GaussianImage++ \citep{li2025gaussianimagepp} & 2D GS                        & -- & $2{,}659$-$3{,}094$ \\
LIG \citep{zhu2025lig}                        & 2D GS                        & -- & $1{,}091$-$1{,}153$ \\
SGI \citep{sgi2026}                           & GS + learned entropy         & $140$-$665$ & $297$-$371$ \\
\textbf{\method{}} (full sphere)              & GS + learned entropy         & $12.3$ & $140$ \\
\textbf{\method{}} (viewport, $1024{\times}768$) & GS + learned entropy      & $8.3$ & $1{,}270$ \\
\bottomrule
\end{tabular}
\caption{Decoding time and rendering speed on one $2048{\times}1024$ ERP
image (one idle RTX~4090). Entropy decoding includes the context networks
of the LIC rows and the rendering-conditioned contexts of \method{};
reconstruction is the synthesis network of the LIC rows and the rendering
of the GS rows. JPEG and JPEG2000 report the full CPU decoding speed, and
-- marks codecs without a learned entropy model. The viewport row decodes
only the streams of the first viewport. Details are given in
App.~\ref{app:viewport}.}
\label{tab:decode_fps}
\end{table}

\paragraph{Qualitative evaluation.}
As shown in Fig.~\ref{fig:qualitative}, \method{} uses the lowest bitrate
among all methods, and compared with GaussianImage and SGI, it preserves
the fine details of the grass and foliage markedly better.

\begin{table}[t]
\centering
\small
\setlength{\tabcolsep}{4pt}
\begin{tabular}{lcc}
\toprule
Variant (default in parentheses) & \multicolumn{2}{c}{$\Delta$rate at equal quality (\%), $+$ is worse} \\
\cmidrule(lr){2-3}
 & WS-PSNR & V-PSNR \\
\midrule
\textbf{Full model} & $0.0$ & $0.0$ \\
\midrule
\multicolumn{3}{l}{\emph{Rate-distortion adaptive primitive selection (Sec.~\ref{sec:field})}} \\
w/o rate-aware opacity weighting ($\omega_{\text{lo}}{=}\omega_{\text{hi}}{=}1$) & $\mathbf{+31.2}$ & $\mathbf{+33.4}$ \\
selection threshold $\tau{=}0.1$ ($\tau{=}0.2$) & $+0.3$ & $+0.4$ \\
selection threshold $\tau{=}0.3$ ($\tau{=}0.2$) & $+0.1$ & $+0.1$ \\
\midrule
\multicolumn{3}{l}{\emph{Rendering-conditioned checkerboard entropy model (Sec.~\ref{sec:entropy})}} \\
w/o \rcc{} & $+5.9$ & $+7.7$ \\
checkerboard on levels $\{5,6\}$ ($\{4,5,6\}$) & $+0.4$ & $+0.3$ \\
STE in the feature rate estimate (uniform noise) & $+2.5$ & $+3.2$ \\
\bottomrule
\end{tabular}
   \caption{Ablation studies of the main components and of the design parameters of each module on the 100 images of the main benchmark: rate difference against the full model at equal quality at $\lambda_R{=}3{\times}10^{-3}$ (positive is worse). The protocol, the variant specifications and the SUN360 results are given in App.~\ref{app:points_abl}.}
\label{tab:ablation}
\end{table}

\subsection{Ablation Studies}
\label{sec:ablation}
We report the WS-PSNR rate difference at equal quality of each variant relative to the full model
(Table~\ref{tab:ablation}; details in App.~\ref{app:points_abl}). For
primitive selection, charging every active primitive its full bitrate
($\omega_{\text{lo}}{=}\omega_{\text{hi}}{=}1$) increases the rate by
$31.2\%$, the largest degradation among all variants, whereas setting
$\tau$ to $0.1$ or $0.3$, which at $b{=}2$ leaves the mask unchanged,
has no measurable effect (at most $0.3\%$). For the entropy model,
replacing \rcc{} with a context-free prior leads to a loss of $5.9\%$,
while restricting the two-pass schedule to levels $\{5,6\}$ has a
negligible impact ($+0.4\%$). Finally, using straight-through rounding instead of additive uniform noise in the feature rate estimate increases the rate by $2.5\%$.

\section{Conclusion}
\label{sec:conclusion}
In this paper, we have presented \method{}, an omnidirectional GS codec
built on hierarchical HEALPix grids. Primitives are anchored at predefined spherical locations, so their coordinates need not be coded and only their existence is signaled. A lightweight entropy model estimates the bitrate of each primitive, and
the existence mask derived from the quantized opacity keeps only the
primitives whose distortion reduction justifies this cost. A single bitstream supports full-sphere,
viewport-dependent, and progressive decoding, and the first viewport
reaches final quality after decoding only $52\%$ of the bitstream.
\method{} outperforms all tested GS codecs, reducing WS-PSNR BD-rate by
$49.6\%$ over GaussianImage++ and $68.6\%$ over SGI.

\bibliographystyle{iclr2027_conference}
\bibliography{references}

\clearpage
\appendix

\section*{Overview of the Appendix}

The appendix is organized as follows.
App.~\ref{app:repro} specifies the implementation details and the evaluation protocol.
App.~\ref{app:rd_ext} gives the full operating points, the pairwise comparisons and the details of the cross-dataset evaluation.
App.~\ref{app:bitdensity} analyzes the bit allocation over the sphere.
App.~\ref{app:viewport} describes the bitstream format and evaluates full-sphere, viewport-dependent and progressive decoding.
App.~\ref{app:points} verifies the bitrate estimate and reports additional ablation and sensitivity results.
App.~\ref{app:l7ceiling} analyzes the resolution limit of the finest level and evaluates a codec with an additional level~7.
App.~\ref{app:qualitative_full} provides additional qualitative results.

\section{Implementation Details and Evaluation Protocol}
\label{app:repro}

This appendix provides the constants, conventions and protocol details referred to in Secs.~\ref{sec:field}--\ref{sec:alloc} and \ref{sec:settings}, including the network architecture and training schedule (App.~\ref{app:repro_model}), the rate and gradient conventions (App.~\ref{app:rate}), and the baselines and metrics (App.~\ref{app:repro_eval}).

\subsection{Architecture and Training}
\label{app:repro_model}

\paragraph{Architecture.}
The spherical hierarchy has $L=7$ primitive levels,
$\ell\in\{0,\dots,6\}$, with $n_{\mathrm{side}}=4\cdot2^{\ell}\in\{4,\dots,256\}$
and $|\Omega^{(\ell)}|=12\,n_{\mathrm{side}}^2$ cells per level
(Sec.~\ref{sec:field}), and a target grid at $n_{\mathrm{side}}=512$ with
$M=3{,}145{,}728$ directions. Each feature
$f^{(\ell)}_i=(f^{(\ell)}_{i,1},\dots,f^{(\ell)}_{i,C})$ has $C=2$ channels
in $[0,1]$, and the opacity $\alpha^{(\ell)}_i$ lies in $[0,1]$ as well.
In the training configuration used throughout the paper, the levels are
coded in three groups: levels $\ell\le2$ are always active, with their
masks and opacities fixed to one and not coded; level 3 takes part in
primitive selection and is coded in a single pass; and levels $\{4,5,6\}$
take part in primitive selection and are coded with the two-pass
checkerboard schedule of Sec.~\ref{sec:entropy}.
The ContextMLP $h_\psi$ has a convolutional structure: the same small
network, with about $3.2$k parameters, is applied to the neighborhood
context of every cell.

\paragraph{Quantization.}
Both attributes are quantized with the uniform quantizer of
Eq.~\eqref{eq:quant} and straight-through gradients
(Sec.~\ref{sec:field}). Each feature channel is quantized with $b=4$ bits,
$\hat f^{(\ell)}_{i,k}=Q_4(f^{(\ell)}_{i,k})$, with the step
$\Delta_f=\Delta_4=1/15$; $\Delta_f$ is also the bin width of the
discretized likelihood in Eq.~\eqref{eq:gmm} and of the uniform noise
added in the rate estimate (Sec.~\ref{sec:field}). The opacity is
quantized with $b=2$ bits,
$\hat\alpha^{(\ell)}_i=Q_2(\alpha^{(\ell)}_i)\in\{0,\tfrac13,\tfrac23,1\}$,
and thresholded at $\tau=0.2$ to give the mask $m^{(\ell)}_i$ of
Eq.~\eqref{eq:mask}.

\paragraph{Rendering and selection constants.}
The weight $w_{v,i}$ of Eq.~\eqref{eq:splat} uses the spherical Gaussian
kernel
\begin{equation}
\label{eq:kernel}
G_{v,i}=\exp\Bigl(-\frac{1-\langle v,u_i\rangle}{(s^{(\ell)})^2}\Bigr),\qquad
s^{(\ell)}=0.5\,\rho^{(\ell)}\,e^{\delta^{(\ell)}},\qquad
\rho^{(\ell)}=\sqrt{4\pi/|\Omega^{(\ell)}|},
\end{equation}
where $u_i$ is the unit vector of the center of cell $i$, $s^{(\ell)}$ is
the isotropic kernel width shared by all primitives of level $\ell$,
$\rho^{(\ell)}$ is the mean angular spacing of the grid, and
$\delta^{(\ell)}$ is learnable. The blending weight of
Eq.~\eqref{eq:blend} is
\begin{equation}
\label{eq:beta}
\beta^{(\ell)}(v)=\min\Bigl(\frac{W^{(\ell)}(v)}{W^{(\ell)}(v)+e^{\gamma^{(\ell)}}},\,0.5\Bigr),\qquad
W^{(\ell)}(v)=\sum_{i\in\mathcal{C}^{(\ell)}(v)}w_{v,i},
\end{equation}
where $W^{(\ell)}(v)$ is the accumulated weight of Eq.~\eqref{eq:splat} and
$\gamma^{(\ell)}$ is a learnable per-level blending parameter. The
remaining constants are listed in Table~\ref{tab:render_constants}. With
$\omega_{\text{lo}}{=}\omega_{\text{hi}}{=}\tau$, the four opacity states
are charged $0$, $\tfrac13$, $\tfrac23$ and $1$ of their feature rate, and
$\omega'{=}1$ on both sides of the threshold. Since the lower branch of
Eq.~\eqref{eq:pw} is reached only at $\hat\alpha{=}0$ in the forward pass,
$\omega_{\text{lo}}$ controls the STE slope of inactive primitives rather
than their forward weight.

\begin{table}[h]
\centering
\small
\begin{tabular}{ll}
\toprule
Constant & Value \\
\midrule
Evaluation grid of Eq.~\eqref{eq:splat} & target grid, $n_{\text{side}}{=}512$, $M{=}3{,}145{,}728$ directions \\
Candidate set $\mathcal{C}^{(\ell)}(v)$ & $K{=}9$ cells (the cell containing $v$ and its 8 neighbors); \\
 & all $|\Omega^{(0)}|{=}192$ primitives at $\ell{=}0$ \\
Denominator floor of Eq.~\eqref{eq:splat} & $\epsilon{=}10^{-8}$ \\
Fill opacity in Eq.~\eqref{eq:gate} & $\alpha_{\mathrm{fill}}{=}0.1$ \\
Selection threshold in Eq.~\eqref{eq:mask} & $\tau{=}0.2$ \\
Opacity weight of Eq.~\eqref{eq:pw} & $\omega_{\text{lo}}{=}\omega_{\text{hi}}{=}\tau$, \ie $\omega(\hat\alpha){=}\hat\alpha$ \\
\bottomrule
\end{tabular}
\caption{Rendering and selection constants.}
\label{tab:render_constants}
\end{table}

\paragraph{Training.}
The training hyperparameters are listed in Table~\ref{tab:training}.

\begin{table}[h]
\centering
\small
\begin{tabular}{ll}
\toprule
Item & Setting \\
\midrule
Optimizer & Adam \citep{kingma2015adam} \\
Learning rate, features & $10^{-2}$ at the finest level, decreasing linearly with the level to $10^{-3}$ at $\ell{=}0$ \\
Learning rate, opacities & $10^{-2}$ \\
Learning rate, entropy model & $3{\times}10^{-3}$ \\
Learning rate, ColorMLP & $3{\times}10^{-3}$ \\
Learning rate, kernel widths & $10^{-3}$ \\
Regularization (Eq.~\eqref{eq:rd}) & $\lambda_s{=}10^{-2}$, $E_{\text{reg}}=\sum_\ell(\delta^{(\ell)})^2+10^{-2}\sum_\ell(\gamma^{(\ell)})^2$ \\
Iterations & 10k per benchmark fit \\
Warm-up & first 6k iterations: $\lambda_R{=}0$ for 3k, then linear increase to its target over the next 3k \\
Fitting time & $12.7$\,min for 10k iterations on one RTX 4090 \\
\bottomrule
\end{tabular}
\caption{Training hyperparameters.}
\label{tab:training}
\end{table}

\subsection{Rate and Gradient Conventions}
\label{app:rate}

\paragraph{Context conventions.}
The context $z^{(\ell)}_i$ of Eq.~\eqref{eq:ctx_main} is fed to the ContextMLP as the concatenation of its three parts and the level index,
\begin{equation}
\label{eq:ctx}
\begin{aligned}
z^{(\ell)}_i&=\bigl[\underbrace{c_{\mathrm{coarse}}(\ell,i)}_{9C}\ \big|\ \underbrace{c_{\mathrm{par}}(\ell,i)}_{C}\ \big|\ \underbrace{c_{\mathrm{even}}(\ell,i)}_{9C}\ \big|\ \ell\bigr],\\
c_{\mathrm{coarse}}(\ell,i)&=\bigl(\bar F^{(\ell-1)}_j\bigr)_{j\in\mathcal{B}_9(i)},\qquad
c_{\mathrm{par}}(\ell,i)=\hat f^{(\ell-1)}_{\mathrm{par}(i)},
\end{aligned}
\end{equation}
which gives a $(19C{+}1)$-dimensional input.
Here $\bar F^{(\ell-1)}_j$ is the coarse-level prediction of Eq.~\eqref{eq:gate}, \ie the mean of the rendering $F^{(\ell-1)}$ over the $4^{\,7-\ell}$ target-grid directions inside cell $j$ (a contiguous NESTED index range), and $\mathrm{par}(i)=\lfloor i/4\rfloor$ is the parent cell of $i$ under the NESTED ordering.
$c_{\mathrm{even}}(\ell,i)$ holds one $C$-dimensional entry per cell $j\in\mathcal{B}_9(i)$: the effective feature $\tilde f^{(\ell)}_j$ of Eq.~\eqref{eq:gate} if cell $j$ has already been decoded when the current pass begins (\ie $\hat f^{(\ell)}_j$ for an active cell and $\bar F^{(\ell-1)}_j$ for an inactive one), and the coarse value $\bar F^{(\ell-1)}_j$ otherwise; in pass~1 and on the single-pass levels, all of its entries are therefore coarse values, and no entry is ever set to zero.
$c_{\mathrm{par}}(\ell,i)$ is set to zero when the parent is inactive.
The two passes of Sec.~\ref{sec:entropy} split the cells of a level by the parity of their NESTED index: pass~1 codes the even cells and pass~2 the odd cells.
Since the lowest bit of a NESTED index is the lowest bit of one face-local coordinate, this split alternates along one axis of each base face: for an odd cell inside a base face, six of its eight neighbors are even and have been decoded in pass~1, while the remaining two, as well as cross-face neighbors of the same parity, take the coarse value $\bar F^{(\ell-1)}_j$, identically at the encoder and the decoder.
The same rule is applied in training, encoding and decoding, which enables the round-trip check of App.~\ref{app:repro_bits}.

\paragraph{Differentiable primitive selection.}
In the charged feature rate $\widetilde R_{\text{feat}}$ of Eq.~\eqref{eq:rfeat_w}, the code length $R^{(\ell)}_i$ is evaluated with the discretized likelihood of Eq.~\eqref{eq:gmm} at the rate-path feature during training, and $\omega\equiv1$ on the always-active levels $\ell\le2$.
Differentiating it gives
\begin{equation}
\label{eq:force}
\frac{\partial \widetilde R_{\text{feat}}}{\partial \alpha^{(\ell)}_i}
\;=\;R^{(\ell)}_i\,\omega'\bigl(\hat\alpha^{(\ell)}_i\bigr),
\end{equation}
where the STE provides $\partial\hat\alpha^{(\ell)}_i/\partial\alpha^{(\ell)}_i{=}1$, and the
context used to predict $R^{(\ell)}_i$ is detached, so that $R^{(\ell)}_i$
acts as a fixed per-primitive cost. The more bits the entropy model predicts for a
primitive, the more strongly its opacity is pushed down.

For an active primitive, Eq.~\eqref{eq:force} is only the rate side of
the gradient on $\alpha$: its $\alpha$ also enters the weighted average
of Eq.~\eqref{eq:splat}, so it receives a gradient from the distortion
term $D$ as well, and the balance between the two decides how the
primitive moves among the four states. A better entropy model therefore
changes not only the code lengths but also which primitives are
active.

\subsection{Baselines and Metrics}
\label{app:repro_eval}

\paragraph{Baselines.} Every GS codec is fitted per image on the same panoramas with its public implementation.
SGI, the only existing GS image codec with a learned entropy model, is fitted per image with its released source code. Its seed budget is varied over $1{,}224$-$10{,}704$ seeds to obtain five operating points covering $0.25$-$1.09$\,bpp, which matches our bitrate range. Its bitrates are the actual bitstream bytes written by its own encoder (anchors, hash grid, MLPs, masks, and positions coded with the MPEG geometry-based point cloud codec (G-PCC)), and its reconstructions are rendered after one encoding-decoding round trip.
LIG, 2DGS-IC \citep{zhang2024_2dgsic} and COIN are fitted with the released code of their authors and evaluated against the original ERP images under the same protocol.
GaussianImage and GaussianImage++ are both fitted with their official released compression configurations: GaussianImage with the two-stage recipe of a $50$k-iteration representation fit followed by $50$k iterations of quantization-aware training, and GaussianImage++ with color normalization enabled and a growth budget above its initial number of primitives, so that its error-driven densification is active.
We restrict the comparison to GS image codecs, with JPEG, JPEG2000 and COIN as reference codecs. JPEG2000, like intra-mode video codecs, still outperforms all compared GS codecs, and we do not claim parity with VVC intra coding.

\paragraph{Metrics.}
\begin{itemize}
    \item \textbf{Resampling.} The HEALPix reconstruction of \method{} is
    first resampled to the $2048{\times}1024$ ERP grid with a Lanczos-3
    kernel, so every method is evaluated on an ERP image. The codecs
    fitted on the ERP plane are thus scored on their own sampling grid,
    whereas \method{} is additionally charged the resampling loss
    (App.~\ref{app:l7ceiling}).
    \item \textbf{WS-PSNR.} WS-PSNR uses the standard latitude weights
    $\varpi(y)=\cos\bigl((y+\tfrac{1}{2}-\tfrac{N_{\mathrm{row}}}{2})\tfrac{\pi}{N_{\mathrm{row}}}\bigr)$
    of the JVET common test conditions \citep{jvet}, where
    $y\in\{0,\dots,N_{\mathrm{row}}-1\}$ indexes the $N_{\mathrm{row}}{=}1024$
    ERP rows.
    \item \textbf{Viewport extraction.} The six viewports (yaw
    $0^\circ/90^\circ/180^\circ/270^\circ$ at the equator, plus zenith and
    nadir) are extracted from every decoded ERP image by bilinear
    \texttt{grid\_sample} projection, identically for all methods. V-PSNR,
    V-SSIM and V-LPIPS are all computed with this projection pipeline.
    \item \textbf{V-SSIM and V-LPIPS.} V-SSIM uses the
    \texttt{pytorch\_msssim} implementation of SSIM \citep{wang2004ssim},
    and V-LPIPS uses the AlexNet LPIPS of \citet{zhang2018unreasonable}
    with inputs scaled to $[-1,1]$; both are computed on the same six
    viewports and averaged over viewports and then over images.
    \item \textbf{BD-rate on V-LPIPS.} BD-rates on V-LPIPS are computed
    after negating the metric, so that the sign convention of
    Table~\ref{tab:bdrate} is the same for all four metrics.
\end{itemize}

\section{Additional Rate-Distortion and Cross-Dataset Results}
\label{app:rd_ext}
\suppressfloats[t]

This appendix supplements the results of Sec.~\ref{sec:rd}. App.~\ref{app:points_op} lists the operating points and the direct pairwise comparisons, and App.~\ref{app:crossdata} gives the protocol and additional statistics of the cross-dataset evaluation.

\subsection{Operating Points and Pairwise BD-Rates}
\label{app:points_op}

\begin{table}[t]
\centering
\small
\setlength{\tabcolsep}{6pt}
\begin{tabular}{cccccc}
\toprule
$\lambda_R$ & bpp & WS-PSNR (dB) & V-PSNR (dB) & V-SSIM & V-LPIPS $\downarrow$ \\
\midrule
$9.0{\times}10^{-3}$  & 0.1448 & 23.83 & 27.61 & 0.7099 & 0.3750 \\
$3.0{\times}10^{-3}$  & 0.2885 & 25.36 & 29.79 & 0.7888 & 0.2522 \\
$2.0{\times}10^{-3}$  & 0.3843 & 26.05 & 30.72 & 0.8202 & 0.2083 \\
$1.4{\times}10^{-3}$  & 0.4931 & 26.67 & 31.58 & 0.8469 & 0.1733 \\
$0.9{\times}10^{-3}$  & 0.6576 & 27.45 & 32.63 & 0.8764 & 0.1372 \\
$4.7{\times}10^{-4}$  & 0.9494 & 28.50 & 34.06 & 0.9103 & 0.0981 \\
$3.0{\times}10^{-4}$  & 1.1675 & 29.09 & 34.86 & 0.9266 & 0.0802 \\
\bottomrule
\end{tabular}
\caption{Operating points of \method{}: mean values over the 100 panoramas, measured end-to-end against the original ERP images (Sec.~\ref{sec:settings}). These are the points of \method{} plotted in Fig.~\ref{fig:rd}. Bitrates are the written bitstream sizes in bits (bytes $\times$ 8) divided by the number of ERP pixels. The decoding measurements of App.~\ref{app:viewport} use the $500$ trained models of the first five points.}
\label{tab:ours_points}
\end{table}

\paragraph{Variation across images.}
At a fixed $\lambda_R$, the bitrate varies by ${\sim}3\times$ across images at the highest bitrate and by ${\sim}4\times$ at the lowest ($0.56$-$1.76$ and $0.07$-$0.26$\,bpp, respectively), so the budget adapts to the image content, which corresponds to the $0.9$-$9.4\%$ range (median $3.5\%$) of level-6 selection ratios at $\lambda_R{=}2{\times}10^{-3}$.

Table~\ref{tab:pairwise} gives the pairwise BD-rates reported in Sec.~\ref{sec:rd}.
Each value is computed on the common quality range of that pair, so the values are not directly comparable with the JPEG-anchored values in Table~\ref{tab:bdrate}. On V-LPIPS, the quality ranges of the two pairs differ enough to reverse the order of SGI and GaussianImage in Table~\ref{tab:bdrate}.

\begin{table}[t]
\centering
\small
\setlength{\tabcolsep}{6pt}
\begin{tabular}{lcccc}
\toprule
Baseline & \multicolumn{4}{c}{\method{} BD-rate (\%) vs.\ baseline $\downarrow$} \\
\cmidrule(lr){2-5}
 & WS-PSNR & V-PSNR & V-SSIM & V-LPIPS \\
\midrule
SGI             & $-68.6$ & $-67.1$ & $-57.7$ & $-66.3$ \\
GaussianImage++ & $-49.6$ & $-45.6$ & $-47.4$ & $-65.1$ \\
GaussianImage   & $-58.2$ & $-56.1$ & $-50.4$ & $-66.7$ \\
LIG             & $-84.2$ & $-82.1$ & $-83.2$ & $-88.6$ \\
\bottomrule
\end{tabular}
\caption{Direct pairwise BD-rates of \method{} against the GS baselines, each computed on the common quality range of that pair.}
\label{tab:pairwise}
\end{table}

\subsection{Cross-Dataset Evaluation}
\label{app:crossdata}

The 100-image benchmark comes from a single source, so the ranking of methods could reflect properties of this dataset rather than of the methods.
We therefore repeat the comparison on all 100 images of the SUN360 \citep{xiao2012sun360} test set provided by \citet{deng2021lau}, an independently collected public 360$^\circ$ dataset distributed as $2048{\times}1024$ JPEG ERP images.
We refit every GS method in this comparison on all images under the protocol of Sec.~\ref{sec:settings}.
We verified that the two datasets do not overlap: a perceptual-hash screening flagged 8 candidate pairs, and a $128{\times}64$ RGB comparison rejected all of them (RMSE $34$-$67$).

\paragraph{Comparison with GS codecs.}
\method{} again outperforms every refitted GS codec at every tested bitrate on all four metrics, with larger margins than on the main benchmark (Fig.~\ref{fig:crossdata}).
At $0.6$\,bpp on V-PSNR, \method{} is $+3.9$\,dB higher than GaussianImage, $+4.1$ than SGI, $+6.9$ than LIG, and $+3.2$ than GaussianImage++, compared with $+3.0$/$+3.7$/$+5.4$/$+2.3$\,dB on the main benchmark. The ranking among the other GS codecs is not fully preserved: GaussianImage and GaussianImage++ swap places on V-SSIM and V-LPIPS (Table~\ref{tab:bdrate}).
Because the SUN360 images are themselves JPEG-compressed, the JPEG anchor re-encodes decoded JPEG content, which may favor it; this can shift the JPEG-anchored SUN360 values in Table~\ref{tab:bdrate}, but not the comparisons among the GS codecs, which do not depend on the anchor.

\paragraph{Statistical significance.}
Since each image yields a paired margin, the margins can be estimated with confidence intervals rather than only by their sign: over the $90$ images on which the curve of \method{} reaches $0.6$\,bpp (the curves of all baselines cover it on every image), the paired V-PSNR margins have standard deviations of $0.9$-$1.4$\,dB, which gives $95\%$ $t$-intervals of $\pm0.2$-$0.3$\,dB (against GaussianImage: $+4.0\pm0.21$\,dB). The $+4.0$\,dB is the mean of the per-image margins, whereas the $+3.9$\,dB above is read from the mean R-D curves; the mean-curve margin over the same 90 images is also $+3.9$\,dB.

\section{Bit Allocation over the Sphere}
\label{app:bitdensity}

In this section, we examine how the feature coding cost of the representation is distributed over the sphere, \ie whether placing primitives on equal-area grid points changes where the bits are spent.

\subsection{Measurement Protocol}
The sphere is divided into $64$ latitude bands of $16$ ERP rows each. Band $n$, with boundary latitudes $\phi_n^{\text{top}}>\phi_n^{\text{bot}}$, covers a solid angle of $A_n=2\pi(\sin\phi_n^{\text{top}}-\sin\phi_n^{\text{bot}})$ steradians, and all methods are divided into bands in the same way.
For \method{}, each coded feature is assigned its exact code length under the CDF table used by the arithmetic coder, at the NESTED position of its primitive. We report these feature bits, which carry the image content; the mask and opacity streams, the header and the MLP weights are not assigned to any band.
For JPEG, the bytes between restart markers inserted after each row of minimum coded units (MCUs) are assigned to their bands; for JPEG2000, the bytes are obtained from the lengths of $2048{\times}16$ tiles in a fixed-quality encoding.
For the learned codecs, cheng2020-attn and MLIC++ with their released MSE weights, the code length $-\log_2 p$ of every latent element is computed from the likelihoods of the entropy model in an evaluation forward pass with rounding: with a stride of $16$, each row of the latent $y$ covers exactly one band, and each row of the hyper-latent $z$, with a stride of $64$, covers four bands, over which its code length is split equally.

For the GS codecs, the bits of each Gaussian are assigned to the band that contains its center. GaussianImage codes the two coordinates of each position with $16$ fixed bits each and its Cholesky and color indices with per-image static histograms, for which we use the ideal code length of each symbol; GaussianImage++ uses $72$ fixed bits per Gaussian in its released code, for the Gaussians that remain after its quantization step; and LIG stores $128$ bits per Gaussian without quantization or entropy coding, with the centers of its half-resolution level scaled to the ERP rows.

\subsection{Bit Allocation over Latitude Bands}
\begin{figure}[t]
\centering
\includegraphics[width=\linewidth]{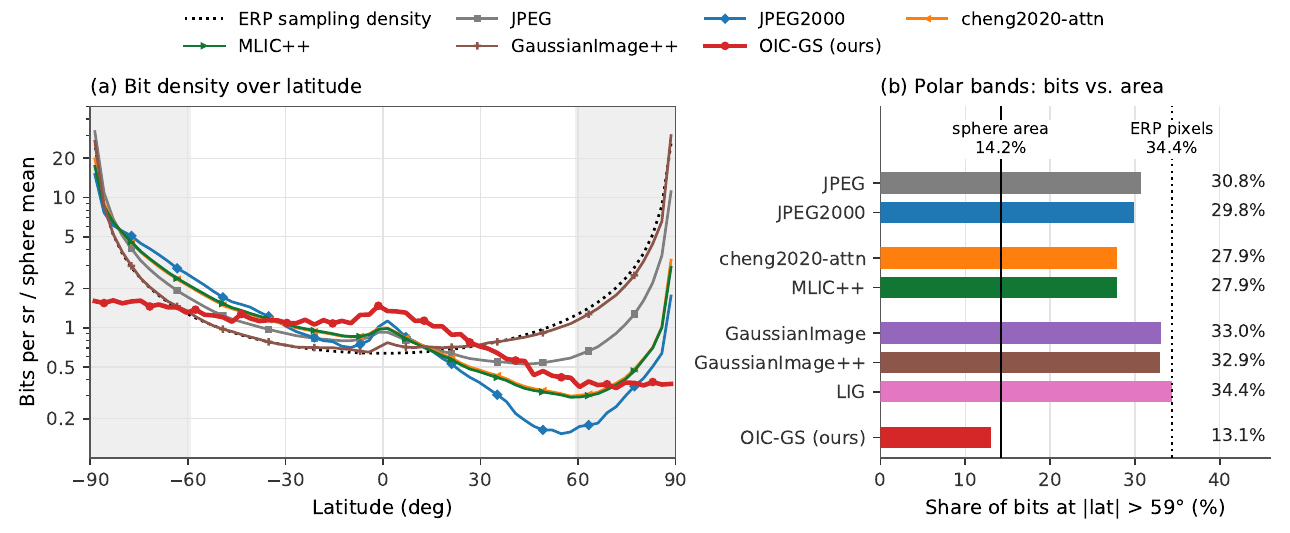}
\caption{Bit allocation over the sphere on the benchmark panoramas at matched bitrates. Bits are assigned to $64$ latitude bands, divided by the solid angle of each band, and normalized by the global mean of each method; for \method{}, the feature bits are shown. \textbf{(a)} The ERP-based codecs follow the sampling density of the projection, while \method{} is distributed substantially more evenly; GaussianImage and LIG are omitted from (a) for readability. \textbf{(b)} The polar bands (the $11$ outermost bands on each side, $|\mathrm{lat}|>59^\circ$) cover $14.2\%$ of the sphere and receive $13.1\%$ of the feature bits of \method{}, close to their area share, compared with $27.9$-$34.4\%$ for the ERP-based codecs, close to the $34.4\%$ ERP pixel share.}
\label{fig:bitdensity}
\end{figure}

As shown in Fig.~\ref{fig:bitdensity}, the equal-area representation removes the concentration of bits at the poles caused by the projection.
The polar share of the ERP-based codecs ($27.9$-$34.4\%$) is about two to two and a half times the $14.2\%$ area share and close to the $34.4\%$ ERP pixel share, \ie their bits follow the number of pixels. This holds in particular for the GS codecs ($32.9$-$34.4\%$), and the bit density of GaussianImage++ closely follows the ERP sampling density (Fig.~\ref{fig:bitdensity}(a)).
The polar share of the feature bits of \method{} ($13.1\%$) is instead close to the $14.2\%$ area share, \ie its feature cost follows the area of the sphere (Fig.~\ref{fig:bitdensity}(b)).
Since the equal-area grid samples the sphere uniformly, the remaining variation of this density (Fig.~\ref{fig:bitdensity}(a)) reflects a redistribution of the feature bits according to the image content.

\section{Bitstream Format and Decoding Modes}
\label{app:viewport}

A panorama viewer can decode the same bitstream in three modes. Path I decodes the whole sphere. Path II decodes only the bytes required by one viewport. Path III performs progressive streaming: it displays the first viewport from part of the file, completes the sphere in the background, and then follows the head motion of the user.
This section first describes the two bitstream formats and the measurement protocol, and then evaluates the three modes. None of them changes the trained representation or affects Eq.~\eqref{eq:rd}.

\subsection{Bitstream Format and Codec Verification}
\label{app:repro_bits}

The default bitstream contains a header, the quantized weights of the two MLPs,
a feature stream for every level coded with arithmetic coding
(global prior at $\ell{=}0$, context-adaptive otherwise), and, for each
selectable level, an existence mask and the $\alpha$ symbols of its active
primitives.
Whereas the feature streams are coded with the learned context model of
Sec.~\ref{sec:entropy}, the mask and the $\alpha$ symbols are written with
simple coders whose probabilities are not learned. Averaged over the 100
panoramas at the lowest and the highest operating point of
Table~\ref{tab:ours_points}, the feature streams take $38$ and $68\%$ of
the file, the mask and opacity streams $30$ and $29\%$ (mask $22$ and
$23\%$, opacity symbols $8$ and $6\%$), and the header and the MLP weights
$32$ and $3\%$.
All reported bitrates of \method{} are byte counts of written files, not
estimates; the differentiable estimates of App.~\ref{app:repro} are used
only inside Eq.~\eqref{eq:rd} and for checkpoint selection.
All R-D results in Sec.~\ref{sec:rd} and the Path~I measurements use full
decoding of this format; the tile-aligned format of App.~\ref{app:tiles}
changes only how the symbols are stored.

Bit-exactness requires the ContextMLP to see a fixed batch shape: otherwise the GPU accumulation order perturbs $\mu$ and $\sigma$ by ${\sim}10^{-5}$, which can move an arithmetic-coding CDF boundary.
The decoder therefore evaluates it over the same full active block as the encoder and passes only the required rows to arithmetic decoding.
Round-trip tests confirm bit-exact recovery of the quantized features and $\alpha$ for every reported configuration, and a pixel-level regression test confirms that viewport-dependent decoding matches full decoding.

The opacity symbols of the active primitives are coded with a static
arithmetic coder: for each selectable level we write the counts of the
three states $\hat\alpha^{(\ell)}_i\in\{\tfrac{1}{3},\tfrac{2}{3},1\}$ as a
header and then code the symbols under that frequency table.
Inactive primitives write no opacity symbol, since the mask already
marks them.

\subsection{Measurement Protocol and Runtime Definitions}
The quality evaluation of Sec.~\ref{sec:settings} uses six square $90^\circ{\times}90^\circ$ viewports, whereas the decoding experiments use viewports with horizontal and vertical fields of view of $90^\circ{\times}60^\circ$ on a $1024{\times}768$ raster to represent a typical viewer workload; the raster size sets the output resolution, while the number of evaluated directions is set by the target grid (App.~\ref{app:vp_render}).
All frame rates of \method{} and the GS baselines are measured on one idle NVIDIA RTX 4090.
We refer to scene 1 at the $\lambda_R{=}2{\times}10^{-3}$ point of Table~\ref{tab:ours_points} as the representative scene.
Bitrates use the number of ERP pixels, and $1$\,KB $=1{,}024$\,B throughout.

We report the frame rates of three rendering modes; all frames stay on the GPU, as in the GS baseline measurements of Table~\ref{tab:decode_fps}. Rendering the full sphere runs at $140$\,FPS. Repeated rendering of a fixed viewport, after its lookup table is built, runs at $1{,}270$\,FPS on average over scenes 1, 7 and 46 and both viewports ($30$ decodes; minimum $1{,}231$) and at $1{,}247$-$1{,}289$\,FPS on the representative scene; $200$ decodes of all $100$ benchmark scenes at the lowest bitrate give $1{,}250$\,FPS. A moving viewport runs at $314$\,FPS on average by re-splatting, and at about $900$\,FPS by looking up a cached sphere. The cached sphere is still slower than a fixed viewport because the two differ in what must be recomputed per frame: a fixed viewport builds its pixel-to-cell lookup table once and reuses it, whereas a moving viewport changes the direction of every pixel in each frame, so the exact NESTED \texttt{ang2pix} has to be evaluated again for all $786{,}432$ pixels of the $1024{\times}768$ frame, more than the $417{,}508$ directions rendered for a fixed viewport. Skipping splatting, blending and the ColorMLP therefore does not bring the frame time below that of a fixed viewport ($1.08$ against $0.79$\,ms; Tables~\ref{tab:frame} and \ref{tab:pan}).
For \method{}, the entropy decoding time in Table~\ref{tab:decode_fps} includes the existence masks and the opacity symbols as well as the feature streams, and all arithmetic decoding runs on the GPU; parsing the stream layout, selecting the streams of a viewport and one-time initialization are excluded.

\paragraph{Frame time breakdown.}
Table~\ref{tab:frame} breaks down the frame of the fixed viewport into its stages.
The viewport path evaluates the $417{,}508$ target-grid directions inside the padded frustum, and a raster lookup maps them to the $1024{\times}768$ pixels. Splatting takes $15\%$ of the GPU time of a frame and blending $3\%$, both restricted to these directions, whereas coarse-level downsampling still operates on the full target grid and takes $19\%$; the ColorMLP takes the largest share, $60\%$. These kernels sum to $0.74$\,ms of the $0.79$\,ms frame; for the rest of the frame the GPU waits for kernel launches. Copying the frame to the host would add $0.85$\,ms and is not part of the benchmark frame rate of $1{,}270$\,FPS, which keeps the frame on the GPU.
Table~\ref{tab:decode_fps} compares the frame rate with those of other codecs.

\begin{table}[t]
\centering
\small
\setlength{\tabcolsep}{6pt}
\begin{tabular}{lrr}
\toprule
 & \multicolumn{2}{c}{Fixed viewport} \\
\cmidrule(lr){2-3}
Stage (per frame) & ms & share \\
\midrule
Level-0 splatting ($192$ primitives)  & $0.046$ & $6.3\%$ \\
Levels 1-6 splatting ($K{=}9$)       & $0.063$ & $8.5\%$ \\
Coarse-level downsampling         & $0.140$ & $19.0\%$ \\
Blending (Eq.~\eqref{eq:blend})   & $0.022$ & $3.0\%$ \\
ColorMLP                          & $0.443$ & $60.2\%$ \\
Raster lookup                     & $0.016$ & $2.2\%$ \\
Other kernels                     & $0.007$ & $0.9\%$ \\
\midrule
Sum of GPU kernel times           & $0.736$ & $100\%$ \\
Frame time       & \multicolumn{2}{r}{$0.787$ ($1{,}270$\,FPS)} \\
\bottomrule
\end{tabular}
\caption{Per-frame time breakdown of the repeated rendering of a fixed viewport (one idle RTX 4090). The viewport rendering evaluates the $417{,}508$ target-grid directions inside the padded $90^\circ{\times}60^\circ$ frustum and maps them to a $1024{\times}768$ frame that stays on the GPU; the host copy is not included. Stage times are GPU kernel times recorded with the PyTorch profiler over $200$ frames, as medians over $15$ models (scenes 1, 7 and 46 $\times$ $5$ bitrates; two viewports, yaw $0^\circ$ and $90^\circ$, per model).}
\label{tab:frame}
\end{table}

\subsection{Tile-Aligned Random Access}
\label{app:tiles}

The default format divides the feature symbols of each coding pass into 32 arithmetic-coding streams with equal numbers of occupied cells.
This format is well suited to parallel decoding on the GPU but poorly suited to skipping, because the stream boundaries are not aligned with spatial tiles, and because the existence mask, the opacities and the coarse levels are each stored as a single block.
We therefore add a tile-aligned format, which can be selected at encoding time without retraining.
The feature streams are divided at NESTED tile boundaries, with $4{,}096$ cells per tile. Levels $4$/$5$/$6$ have $12$/$48$/$192$ tiles, each containing one feature segment per coding pass, which gives $96$/$384$ tile streams on levels $5$/$6$.
The existence mask and the opacities are likewise coded per tile, all segment sizes are stored in the header, and the decoder detects the format from a header flag.
The mask, opacity and feature segments of every tile are then contiguous byte ranges that can be skipped independently.
Similar to the precinct size of JPEG2000, the tile size is fixed at encoding time. The difference is that the tiles are equal-area HEALPix cells, so a polar viewport is not expanded into a full-width band of tiles (our decoding measurements use equatorial viewports).

\begin{table}[t]
\centering
\small
\begin{tabular}{lccc}
\toprule
Format & File size & $\Delta$ size & Saved (yaw $0^\circ$/$90^\circ$) \\
\midrule
32 equal streams per pass (default)  & $116{,}089$\,B ($0.4429$\,bpp) & -- & -- \\
Tile-aligned ($4{,}096$ cells/tile)  & $117{,}352$\,B ($0.4477$\,bpp) & $+1.1\%$ & $50.0\%$ / $39.4\%$ \\
\bottomrule
\end{tabular}
\caption{Two storage formats of the same symbols on the representative scene ($90^\circ{\times}60^\circ$ viewports). The tile-aligned format increases the file size by $1.1\%$ ($1.2\%$ benchmark mean) and allows $39$-$50\%$ of the file to be skipped for one viewport; the default format has no tile boundaries to skip.}
\label{tab:layouts}
\end{table}

\subsection{Full-Sphere and Viewport-Dependent Decoding}
\label{app:vp_render}

As described in Sec.~\ref{sec:field}, a viewport is rendered by evaluating Eqs.~\eqref{eq:gate}--\eqref{eq:color} only for its pixel directions. In our implementation, each pixel direction is mapped to its cell on the target grid by a precomputed lookup table, and the equations are evaluated once per visible target-grid cell, without an ERP intermediate. The target grid has a mean angular spacing of about $0.11^\circ$, comparable to the pixel pitch at the center of the $1024{\times}768$ viewport ($0.09$-$0.11^\circ$), so the viewport is rendered at approximately display resolution.
Given a padded viewing frustum, we select the visible cells of the finest grid and propagate visibility to coarser levels of the NESTED hierarchy by taking the union over children, which gives one visibility mask per level, and restrict the neighbor tables of Eqs.~\eqref{eq:splat}--\eqref{eq:blend} to the visible rows.
A $90^\circ{\times}60^\circ$ frustum with $3^\circ$ padding covers about $13\%$ of the target directions ($417{,}508$ of $3{,}145{,}728$), but the skipped fraction of the file is much smaller, because the coarse levels and the context dependencies are shared.
The visibility culling is bit-exact with full rendering inside the viewport. The viewport-dependent and progressive measurements below use the tile-aligned format, while Path~I decodes the default format.

\paragraph{Path I: full-sphere decoding.}
\label{app:path_full}

Full decoding recovers the quantized features and opacities of the encoder bit-exactly, and reconstructs the sphere of the representative scene at $27.7133$\,dB native PSNR (PSNR on the HEALPix representation).
Full-sphere rendering runs at $140$\,FPS, and repeated rendering of a fixed viewport from the decoded representation reaches $1{,}247$-$1{,}289$\,FPS on the representative scene; Table~\ref{tab:frame} gives the breakdown of the viewport frame.

\paragraph{Path II: viewport-dependent decoding.}
\label{app:path_pruned}

A two-pass procedure first parses the stream boundaries, then extends the viewport with its ring, parent and two-pass dependencies, and decodes only the streams that intersect it; skipped primitives can only affect pixels outside the viewport.
This is verified on the two viewports of the representative scene and on all $500$ benchmark models with two viewports each: inside the viewport, every decode is bit-exact with full decoding.

\subsection{Progressive Streaming and Head Rotation}
\label{app:path_stream}

Path III uses the tile-aligned format for streaming.
A session has three phases: in phase $T_0$, the tiles of the first viewport are received and decoded; in phase $T_1$, the remaining tiles complete the sphere in the background; and in phase $T_2$, every head rotation only requires rendering.
Table~\ref{tab:timeline} measures the whole process on scenes 1, 7 and 46, one fresh process per session, and Table~\ref{tab:pan} compares the two strategies for head rotation on the representative scene over a $180$-frame pan.
In phase $T_2$, the two strategies differ in what is computed per frame.
The re-splatting strategy renders every frame from the primitives: it
runs the splatting of Eq.~\eqref{eq:splat}, the blending of
Eq.~\eqref{eq:blend} and the ColorMLP on the directions of the current
viewport. To avoid recomputing visibility at every frame, it keeps a
visible set with a margin around the viewport and updates it only when
the viewport leaves this margin, which happens four times during the
$180$-frame pan. The cached strategy instead renders the full sphere once
on the target grid after phase $T_1$ ($7.1$\,ms at $140$\,FPS) and
stores the RGB color of every target direction. Each frame then only maps
the direction of every viewport pixel to its target-grid cell and reads
the cached color, so splatting, blending and the ColorMLP are skipped
entirely, at the cost of one full-sphere rendering and a color buffer for
the whole sphere. Since the re-splatting strategy evaluates the same
target-grid directions and uses the same direction-to-cell lookup, the
two strategies produce bit-identical frames. Both use an exact GPU implementation of the NESTED \texttt{ang2pix} function (zero mismatches against \texttt{healpy} \citep{zonca2019healpy} over $2{\times}10^5$ directions) and produce identical PSNR along the trajectory in every session; a per-frame CPU lookup, which rebuilds the lookup table every frame, is far slower and is not reported.

\begin{table}[t]
\centering
\footnotesize
\setlength{\tabcolsep}{3pt}
\begin{tabular}{lccccc}
\toprule
$\lambda_R$ & First view (bpp) & Deferred & $T_1$ (KB) & Unchanged & $T_2$ (FPS) \\
\midrule
$0.9{\times}10^{-3}$  & $0.338$ & $48\%$ & $82$  & $3/3$ & $314$ / $891$ \\
\bottomrule
\end{tabular}
\caption{Progressive streaming: the models of scenes 1, 7 and 46 at the $\lambda_R{=}0.9{\times}10^{-3}$ point of Table~\ref{tab:ours_points} played end-to-end in fresh processes ($3$ sessions; first viewport at yaw $0^\circ$, followed by a $180$-frame pan to $90^\circ$). ``$T_1$ (KB)'' is the data size of phase $T_1$, and ``$T_2$ (FPS)'' is the rendering speed of phase $T_2$. ``Deferred'' is the fraction of the file not yet received when the first viewport is displayed in each session. ``Unchanged'' counts the sessions whose first-viewport pixels are bit-identical after completion; $T_2$ gives the re-splatting/cached speeds of Table~\ref{tab:pan}.}
\label{tab:timeline}
\end{table}

\begin{table}[t]
\centering
\small
\begin{tabular}{lcccl}
\toprule
Strategy & ms/frame & FPS & GT-PSNR & Notes \\
\midrule
Re-splatting, visible set with margin & $3.22$ & $310$  & $27.994$ & $4$ updates of the visible set \\
Cached sphere, lookup only      & $1.08$ & $\mathbf{923}$ & $27.994$ & bit-identical to re-splatting \\
\bottomrule
\end{tabular}
\caption{Head-rotation speed on the representative scene ($180$ frames from yaw $0^\circ$ to $90^\circ$, $1024{\times}768$ at $90^\circ{\times}60^\circ$, idle RTX 4090). FPS is computed from the unrounded mean frame time. Both strategies are measured in a single run that decodes once and then pans; over the sessions of Table~\ref{tab:timeline}, they run at $314$/$891$\,FPS on average. GT-PSNR, the mean viewport PSNR along the trajectory against the ground-truth projection, is identical for all strategies, so the acceleration does not affect quality.}
\label{tab:pan}
\end{table}

\begin{figure}[t]
\centering
\includegraphics[width=\linewidth]{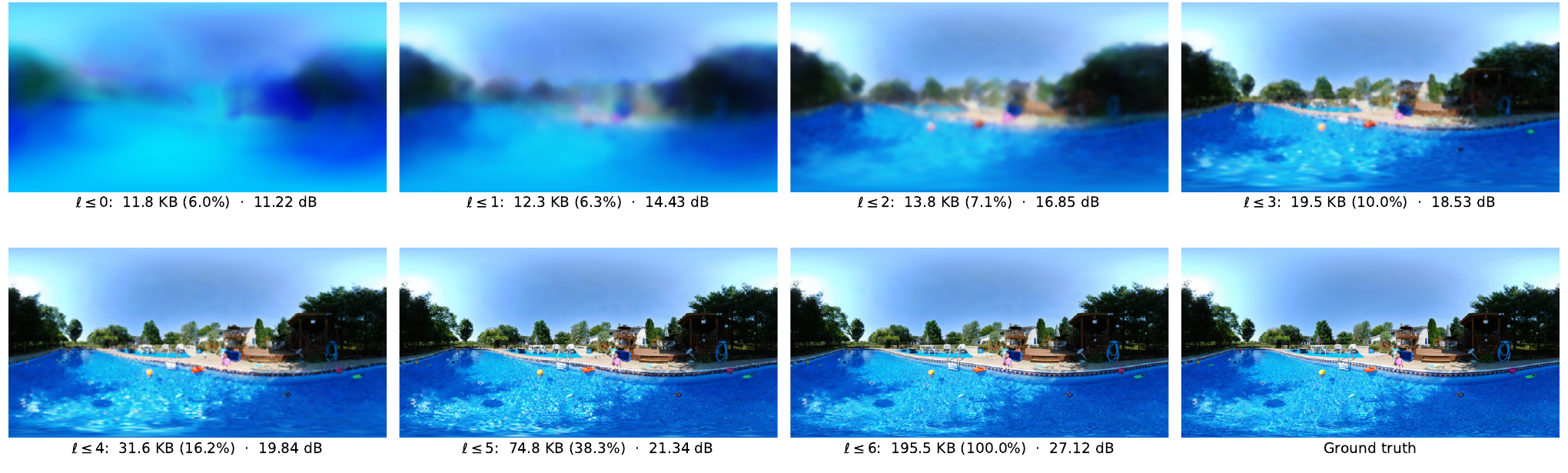}
\caption{Progressive decoding by level, from the written bitstream of scene 1 at the $\lambda_R{=}0.9{\times}10^{-3}$ operating point ($195.5$\,KB; Lanczos-3 resampling to ERP). Each panel shows the full sphere reconstructed only from the levels $\ell\le L'$, labeled with the bytes of this prefix (share of the file) and the end-to-end WS-PSNR. Because of the coarse-to-fine order of Sec.~\ref{sec:entropy}, every prefix gives a complete sphere with less detail: a recognizable preview requires the first $10\%$ of the file (levels $\ell\le3$), and the final prefix is bit-exact with full decoding.}
\label{fig:prefix}
\end{figure}

\subsection{Deployment Considerations}
\label{app:deploy}

\paragraph{Bandwidth and completion.}
Path II displays a bit-exact viewport at final quality from the tiles that the viewport requires. Path III uses sessions with a fixed initial yaw and displays the first viewport from $52\%$ of the file; after the remaining tiles arrive, the sphere is completed without changing any pixel on screen (Table~\ref{tab:timeline}). Level-aligned prefixes serve a different purpose: they provide a coarse preview of the full sphere from the first $10\%$ of the file (Fig.~\ref{fig:prefix}).

\paragraph{Interaction and infrastructure.}
Both GPU strategies for head rotation stay within the $11.1$\,ms frame budget of a $90$\,Hz display (Table~\ref{tab:pan}). Since tile sizes and offsets are stored explicitly, a static file supports viewport-adaptive HTTP range requests without server-side transcoding or per-client bitstreams; the storage overhead of the tile-aligned format is $1.2\%$ on average (Table~\ref{tab:layouts}).

\section{Additional Ablation Studies and Bitrate Estimate Analysis}
\label{app:points}
\suppressfloats[t]

This appendix extends Sec.~\ref{sec:ablation}.
App.~\ref{app:points_bits} compares the bitrate estimate used for selection with the actually written bytes. App.~\ref{app:points_abl} gives a detailed analysis of the ablation variants of Table~\ref{tab:ablation} and the sensitivity to the design constants.

\subsection{Estimated Bitrate versus Written File}
\label{app:points_bits}

We compare the bitrate estimate used in training with the actually written bytes over the encodings of Table~\ref{tab:ours_points}.

\paragraph{Accuracy of the bitrate estimate.}
Let $R_{\text{feat}}$ denote the masked feature rate of Eq.~\eqref{eq:rate_main}, \ie $\sum_{\ell}\sum_{i}m^{(\ell)}_i R^{(\ell)}_i$ with the code lengths $R^{(\ell)}_i$ of Eq.~\eqref{eq:rate_main} evaluated at the quantized $\hat f^{(\ell)}_i$ of the symbols actually coded. It is obtained from $\widetilde R_{\text{feat}}$ (Eq.~\eqref{eq:rfeat_w}) by setting $\omega{\equiv}1$ and excluding the inactive primitives, whereas the $\omega$-weighted $\widetilde R_{\text{feat}}$ is used only in the training objective.
The written feature bytes exceed $R_{\text{feat}}$ by only $1.9\%$ on average (median $0.7\%$), which can be attributed to the termination overhead of the $32$ parallel streams.

\subsection{Ablation Variants and Design Constants}
\label{app:points_abl}

All variants of Table~\ref{tab:ablation} are retrained from scratch on all 100 images of the main benchmark at the operating point $\lambda_R{=}3{\times}10^{-3}$ of Table~\ref{tab:ours_points}, with the schedule of Table~\ref{tab:training}; to verify that the conclusions do not depend on the dataset, they are also retrained on all 100 SUN360 images of App.~\ref{app:crossdata}.
Each variant is compared with the full model at equal quality: its mean operating point is projected onto the mean rate-distortion curve of the full model on the same dataset (seven operating points; Table~\ref{tab:ours_points} for the main benchmark), interpolated with PCHIP in quality against $\log_{10}$ bpp, and Table~\ref{tab:ablation} reports the relative rate difference at the quality of the variant.
Since this operating point is bracketed by the curve of the full model, the WS-PSNR and V-PSNR of every variant lie within the quality range of the full model on every image of both datasets, so no extrapolation is needed.
On SUN360, the variants give $+29.0\%$/$+28.1\%$ (WS-PSNR/V-PSNR) without rate-aware opacity weighting, $+6.5\%$/$+7.9\%$ without \rcc{}, $+2.4\%$/$+3.3\%$ with STE instead of uniform noise in the feature rate estimate, and between $-0.6\%$ and $+0.5\%$ for the design constants, which reproduces the ranking on the main benchmark.

\paragraph{Rate-aware opacity weighting ($+31.2\%$).}
This variant sets $\omega_{\text{lo}}{=}\omega_{\text{hi}}{=}1$ in Eq.~\eqref{eq:pw}, \ie $\omega(\hat\alpha)=\min(\hat\alpha/\tau,1)$. At $b{=}2$, its forward value is the mask $m^{(\ell)}_i=\mathbf{1}[\hat\alpha^{(\ell)}_i>\tau]$ of Eq.~\eqref{eq:mask}, so it removes the rate slope above $\tau$ and charges every active primitive its full bitrate, while the straight-through slope $1/\tau$ still pushes inactive primitives down.
The variant is worse than the full model on $96$ of the $100$ images of the main benchmark, and the effect is visible in the selection: at the same $\lambda_R$, it keeps $6.1\times$ as many active primitives ($288{,}576$ against $47{,}031$ on average) and spends $0.66$ instead of $0.29$\,bpp.
The deviation between its written and estimated feature bytes is $2.5\%$, similar to that of the full model (App.~\ref{app:points_bits}), so this variant is itself a well-calibrated codec.
Another possible variant, $\omega{\equiv}1$, would also charge the objective for the bits of inactive primitives, which are never written by the coder. Since the features of inactive primitives receive no distortion gradient, this variant would only push them toward cheaper values that are never transmitted.
We therefore use the former variant as the ablation of \rpg{}.

\paragraph{\rcc{} ($+5.9\%$).}
This variant sets all context inputs of the ContextMLP to zero, so that the network reduces to one learned mixture per level, and it is retrained end to end, so the representation re-adapts to the weaker prior.
It is worse than the full model on $98$ of the $100$ images of the main benchmark.
The re-adaptation is visible in the selection: without context, the variant selects $0.84\times$ as many primitives, because a prior without context raises the predicted cost of an average primitive, so Eq.~\eqref{eq:force} pushes all opacities down more strongly at a fixed $\lambda_R$.
\rcc{} therefore lowers the predicted cost of an average primitive, which allows more primitives to be kept at the same $\lambda_R$.

\paragraph{Uniform noise in the feature rate estimate ($+2.5\%$).}
With straight-through rounding instead of uniform noise in the feature rate estimate, the likelihood of Eq.~\eqref{eq:gmm} is evaluated at the rounded features and still estimates the bitrate of the coded symbols; only its gradient changes, so the quality is essentially unaffected (within $0.02$\,dB WS-PSNR on the main benchmark) and the loss appears in the bitrate. The variant is worse than the full model on $96$ of the $100$ images of the main benchmark.

\paragraph{Design constants.}
The two-pass schedule has no measurable effect (Table~\ref{tab:ablation}): removing level $4$ from it changes the rate by $+0.4\%$ to $+0.3\%$, which is of the order of the run-to-run variation of the training ($0.48\%$ in bpp, $0.027$\,dB in native PSNR over repeated identical runs), as expected since level 4 contains less than $5\%$ of the cells coded with the two-pass schedule.
\section{Resolution Limit and the L7 Extension}
\label{app:l7ceiling}

The main model ends at level 6 ($n_{\text{side}}{=}256$), one resolution level below its target and resampling grid at $n_{\text{side}}{=}512$.
In this appendix, L6 denotes this default configuration ($L{=}7$ levels, $\ell\in\{0,\dots,6\}$), and L7 denotes the configuration with an additional level $\ell{=}7$ at $n_{\text{side}}{=}512$ ($L{=}8$ levels).
App.~\ref{app:l7capacity} measures the capacity of the two configurations without a rate term, and App.~\ref{app:l7codec} evaluates the L7 codec at matched bitrates.

\subsection{Capacity Analysis without Rate}
\label{app:l7capacity}

\begin{figure*}[t]
\centering
\includegraphics[width=\textwidth]{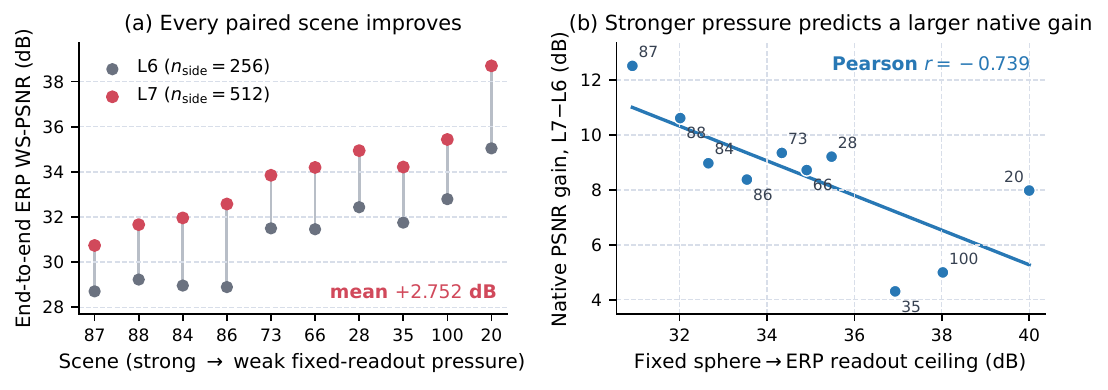}
\caption{Resolution analysis. \textbf{Left:} the finest level at $n_{\text{side}}{=}512$ improves the end-to-end ERP WS-PSNR for every selected image ($+2.752$\,dB on average; range $+2.035$-$+3.682$\,dB). \textbf{Right:} images with lower round-trip fidelity of the reference ERP$\rightarrow$HEALPix$\rightarrow$ERP path, which indicates stronger resampling loss and more high-frequency content, obtain a larger native L7 gain (Pearson correlation $-0.739$). The native gain is also correlated with the global and nadir high-frequency energy ($+0.642$ and $+0.721$), and is $+10.480$\,dB on average at the nadir against $+8.022$\,dB at the equator.}
\label{fig:l7ceiling}
\end{figure*}

\paragraph{Setup.}
To select test images without using the results of our method, we rank all 100 images by the WS-PSNR of the model-independent ERP $\rightarrow$ HEALPix $\rightarrow$ Lanczos-3 ERP round trip, divide the ranking into ten bins, and take the middle image of each bin (scenes 87, 88, 84, 86, 73, 66, 28, 35, 100 and 20).
For each scene, we fit paired L6 and L7 models for 10k iterations with the same fixed seed and optimization settings.
L6 contains 1,048,512 primitives up to $n_{\text{side}}{=}256$; L7 adds the 3,145,728 cells of the target grid, for 4,194,240 in total.
Both fits use $\lambda_R{=}0$ and continuous features, keep all primitives active, and select the best checkpoint every 250 iterations.
This is therefore a capacity analysis with a fixed number of primitives, not a comparison at matched bitrates; Table~\ref{tab:l7ceiling} and Fig.~\ref{fig:l7ceiling} report the results for each scene.

\begin{table*}[t]
\centering
\small
\setlength{\tabcolsep}{4.2pt}
\begin{tabular}{lrrrrrrr}
\toprule
& Reference round-trip & \multicolumn{3}{c}{Native sphere PSNR (dB)} & \multicolumn{3}{c}{End-to-end ERP WS-PSNR (dB)} \\
\cmidrule(lr){2-2}\cmidrule(lr){3-5}\cmidrule(lr){6-8}
Scene & fidelity & L6 & L7 & $\Delta$ & L6 & L7 & $\Delta$ \\
\midrule
87  & 30.921 & 32.952 & 45.466 & $+12.514$ & 28.702 & 30.737 & $+2.035$ \\
88  & 32.016 & 33.349 & 43.963 & $+10.614$ & 29.225 & 31.660 & $+2.435$ \\
84  & 32.659 & 32.515 & 41.484 &  $+8.969$ & 28.959 & 31.959 & $+3.000$ \\
86  & 33.541 & 32.290 & 40.666 &  $+8.375$ & 28.892 & 32.574 & $+3.682$ \\
73  & 34.341 & 35.009 & 44.354 &  $+9.345$ & 31.496 & 33.848 & $+2.352$ \\
66  & 34.910 & 34.702 & 43.424 &  $+8.722$ & 31.456 & 34.196 & $+2.740$ \\
28  & 35.479 & 35.584 & 44.792 &  $+9.209$ & 32.432 & 34.941 & $+2.508$ \\
35  & 36.934 & 34.158 & 38.465 &  $+4.307$ & 31.749 & 34.214 & $+2.465$ \\
100 & 38.024 & 35.226 & 40.222 &  $+4.996$ & 32.790 & 35.436 & $+2.646$ \\
20  & 40.000 & 37.823 & 45.799 &  $+7.975$ & 35.042 & 38.699 & $+3.657$ \\
\midrule
Mean & 34.882 & 34.361 & 42.863 & $\mathbf{+8.502}$ & 31.074 & 33.826 & $\mathbf{+2.752}$ \\
\bottomrule
\end{tabular}
\caption{Paired L6/L7 resolution analysis. ``Reference round-trip fidelity'' is the WS-PSNR of the original ERP image after the reference ERP$\rightarrow$HEALPix$\rightarrow$ERP round trip; it is used to stratify the image selection and as the reference score in Fig.~\ref{fig:l7ceiling}, and it is not a mathematical upper bound. Both the native and the end-to-end scores improve on all $10/10$ scenes. These fits use only the distortion loss, continuous features and all primitives active, and are therefore not codec operating points or results at matched bitrates.}
\label{tab:l7ceiling}
\end{table*}

\paragraph{Findings.}
The additional level raises the native sphere PSNR by $+8.502$\,dB and the end-to-end WS-PSNR by $+2.752$\,dB on average, on all $10/10$ scenes, so level 6 is an empirical capacity limit under this setting.
The end-to-end gain is smaller because the ERP score includes the two resamplings between the ERP grid and the sphere: the reference round trip, which involves no codec, reaches only $34.882$\,dB on these scenes, and the gap of the fits to it decreases from $3.808$ to $1.056$\,dB.
Codecs fitted on the ERP plane never perform these resamplings, so the ERP-based scores of Sec.~\ref{sec:rd} charge \method{} an additional resampling loss (App.~\ref{app:repro_eval}); on the main benchmark, the operating points of Table~\ref{tab:ours_points} score $1.9$-$2.8$\,dB higher in the native domain than end-to-end.
Whether the additional level is worth its bitrate once the rate term, quantization and selection are enabled is evaluated next.

\subsection{Rate-Distortion Evaluation with L7}
\label{app:l7codec}

\paragraph{Setup.}
We retrain the full codec with the additional level on a 20-image subset of the main benchmark (scenes 20, 21, 25, 34, 42, 46, 48, 51, 52, 62, 67, 71, 75, 80, 86, 87, 89, 95, 98 and 100), changing only the number of levels and the two-pass schedule ($\ell\in\{4,5,6,7\}$); all other hyper-parameters follow App.~\ref{app:repro}, and the target and resampling grid remains $n_{\text{side}}{=}512$.
Both configurations are evaluated at $\lambda_R\in\{2.0,1.4,0.9\}{\times}10^{-3}$ with the rate term, quantization and \rpg{} enabled, so the additional level must justify its own bitrate.
All bitrates are byte counts of written bitstreams, and the round-trip properties of App.~\ref{app:repro_bits} still hold: the decoded PSNR matches the encoder within $0.0014$\,dB and re-encoding gives identical bytes on $60/60$ runs.
Fitting L7 takes $18.0$ against $12.7$\,min per image, and its run-to-run variation is larger ($0.05$-$0.15$ against $0.027$\,dB; App.~\ref{app:points_abl}), so we keep L6 as the default and report L7 as a scalability result: the same selection mechanism extends to a finer level without any other change.

\paragraph{Rate-distortion performance.}
The L7 codec reduces the WS-PSNR BD-rate by $7.7\%$ relative to L6 ($+0.24$\,dB at matched bitrate) and the V-PSNR BD-rate by $8.2\%$, on all $20/20$ images with a defined BD-rate (Table~\ref{tab:l7rd}).

\begin{table}[t]
\centering
\small
\begin{tabular}{lccc}
\toprule
Metric & BD-rate (\%) & $\Delta$ at matched bitrate & Negative on \\
\midrule
WS-PSNR & $\mathbf{-7.7}$ & $+0.24$\,dB & $20/20$ \\
V-PSNR & $-8.2$ & $+0.35$\,dB & $20/20$ \\
native sphere PSNR & $-11.7$ & $+0.46$\,dB & $20/20$ \\
\bottomrule
\end{tabular}
\caption{The L7 codec against the L6 configuration on the 20-image subset of App.~\ref{app:l7codec}, at the three operating points $\lambda_R\in\{2.0,1.4,0.9\}{\times}10^{-3}$, computed with the estimator of Table~\ref{tab:bdrate} on the common quality range of each pair. A negative value means fewer bits at equal quality; ``Negative on'' counts the images with a defined BD-rate. Native sphere PSNR is measured on the HEALPix representation and excludes the effect of resampling.}
\label{tab:l7rd}
\end{table}

\begin{figure}[t]
\centering
\includegraphics[width=0.78\linewidth]{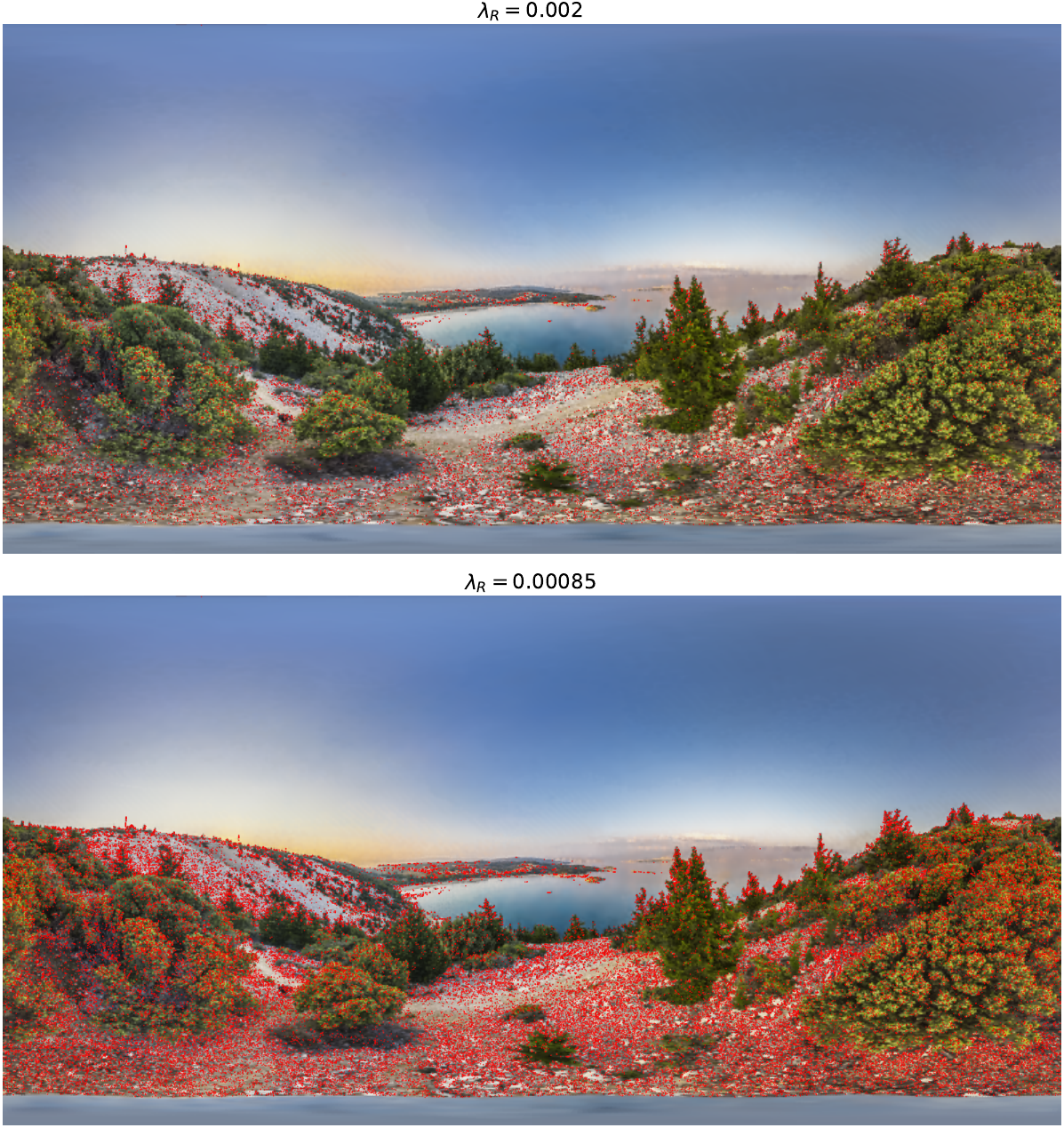}
\caption{Locations of active primitives of the additional level on scene 87: active level-7 cells are shown in red over the ground truth. They concentrate on foliage, rocky ground and the shoreline, and are absent from the sky and open water; decreasing $\lambda_R$ (bottom) selects more of them without changing where they are located.}
\label{fig:l7survivors}
\end{figure}

\paragraph{Selection on the additional level.}
The median image selects only $0.14$-$0.56\%$ of the additional level across the three bitrates (per-image range $0.02$-$1.87\%$; Table~\ref{tab:l7survival}).
Level 6 loses $28$-$56\%$ of its active primitives while levels 4-5 keep slightly more, and the total number of active primitives is unchanged at the lowest bitrate and $9$-$15\%$ lower at the two highest, so \rpg{} redistributes primitives across levels rather than simply adding more.
The selection is not degenerate: the median selection ratio of level 7 increases by $4.0\times$ as $\lambda_R$ decreases by $2.2\times$; all cells start above $\tau$ (mean opacity $0.616$), and $45\%$ of level 7 is still active after the 3k warm-up iterations without the rate term, before the rate term selects among them; and the active primitives concentrate on high-frequency content, with a ground-truth Sobel gradient of $2.4$-$4.7\times$ the solid-angle-weighted image mean and $65$-$88\%$ of them in the top gradient quartile (against $1.9$-$3.2\times$ for level 6).
Fig.~\ref{fig:l7survivors} shows their locations on scene 87; the polar caps receive more than their $13.4\%$ solid-angle share of them, because the nadir often contains textured ground, where the native L7 gain is also the largest (Fig.~\ref{fig:l7ceiling}).

\begin{table}[t]
\centering
\small
\setlength{\tabcolsep}{4.5pt}
\begin{tabular}{lccccc}
\toprule
& \multicolumn{1}{c}{Level 7} & \multicolumn{3}{c}{Selection ratio, L6 config $\rightarrow$ L7 config} & Active primitives \\
\cmidrule(lr){2-2}\cmidrule(lr){3-5}\cmidrule(lr){6-6}
$\lambda_R$ & median (range) & Level 4 & Level 5 & Level 6 & L6 $\rightarrow$ L7 \\
\midrule
$2.0{\times}10^{-3}$   & $0.14\%$ ($0.02$-$0.60$) & $71.9\rightarrow76.1\%$ & $34.0\rightarrow42.3\%$ & $8.7\rightarrow6.3\%$ & $186$K $\rightarrow$ $191$K \\
$1.4{\times}10^{-3}$   & $0.25\%$ ($0.04$-$1.02$) & $73.6\rightarrow77.8\%$ & $37.2\rightarrow46.6\%$ & $11.8\rightarrow5.2\%$ & $217$K $\rightarrow$ $197$K \\
$0.9{\times}10^{-3}$   & $0.56\%$ ($0.11$-$1.87$) & $75.4\rightarrow78.6\%$ & $45.8\rightarrow49.9\%$ & $18.2\rightarrow8.4\%$ & $285$K $\rightarrow$ $241$K \\
\bottomrule
\end{tabular}
\caption{Selection ratios after training, once the inactive primitives are removed, on the 20-image subset of App.~\ref{app:l7codec} at the three operating points $\lambda_R\in\{2.0,1.4,0.9\}{\times}10^{-3}$. Level 7 contains $3{,}145{,}728$ cells; the median column gives the per-image median selection ratio (range in parentheses), and the remaining columns compare the same level between the two configurations at the same $\lambda_R$.}
\label{tab:l7survival}
\end{table}

\clearpage
\section{Additional Qualitative Results}
\label{app:qualitative_full}

\begin{figure}[!htbp]
\centering
\includegraphics[width=\textwidth]{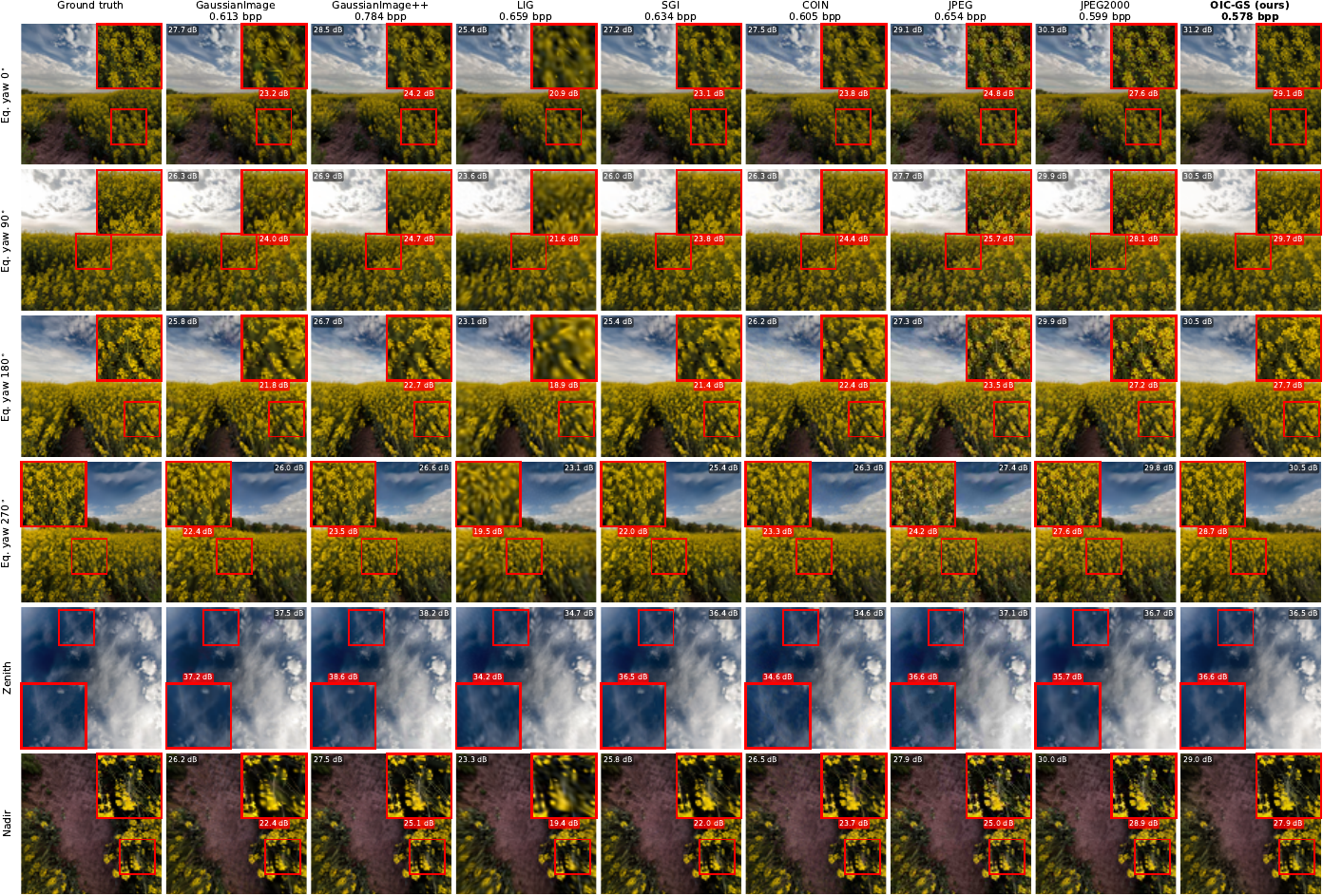}
\caption{Additional qualitative results (part 1 of 3): scene 55 in all six viewing directions, under the protocol described in the text. Across the five scenes, \method{} achieves the highest viewport PSNR among the GS codecs and COIN in $25$ of the $30$ panels, and the highest patch PSNR in $27$. On this scene, the only exception is the zenith, where GaussianImage++ and GaussianImage exceed \method{} by $1.7$ and $1.0$\,dB at $0.784$ and $0.613$\,bpp, compared with $0.578$\,bpp for \method{}; the other four exceptions are shown in Fig.~\ref{fig:qualgrid2}. \method{} matches or exceeds JPEG2000 in $11$ of the $12$ equatorial panels of scenes 7, 32 and 55, while JPEG2000 remains ahead at the nadir of every scene.}
\label{fig:qualgrid1}
\end{figure}

\begin{figure}[t]
\centering
\includegraphics[width=0.92\textwidth]{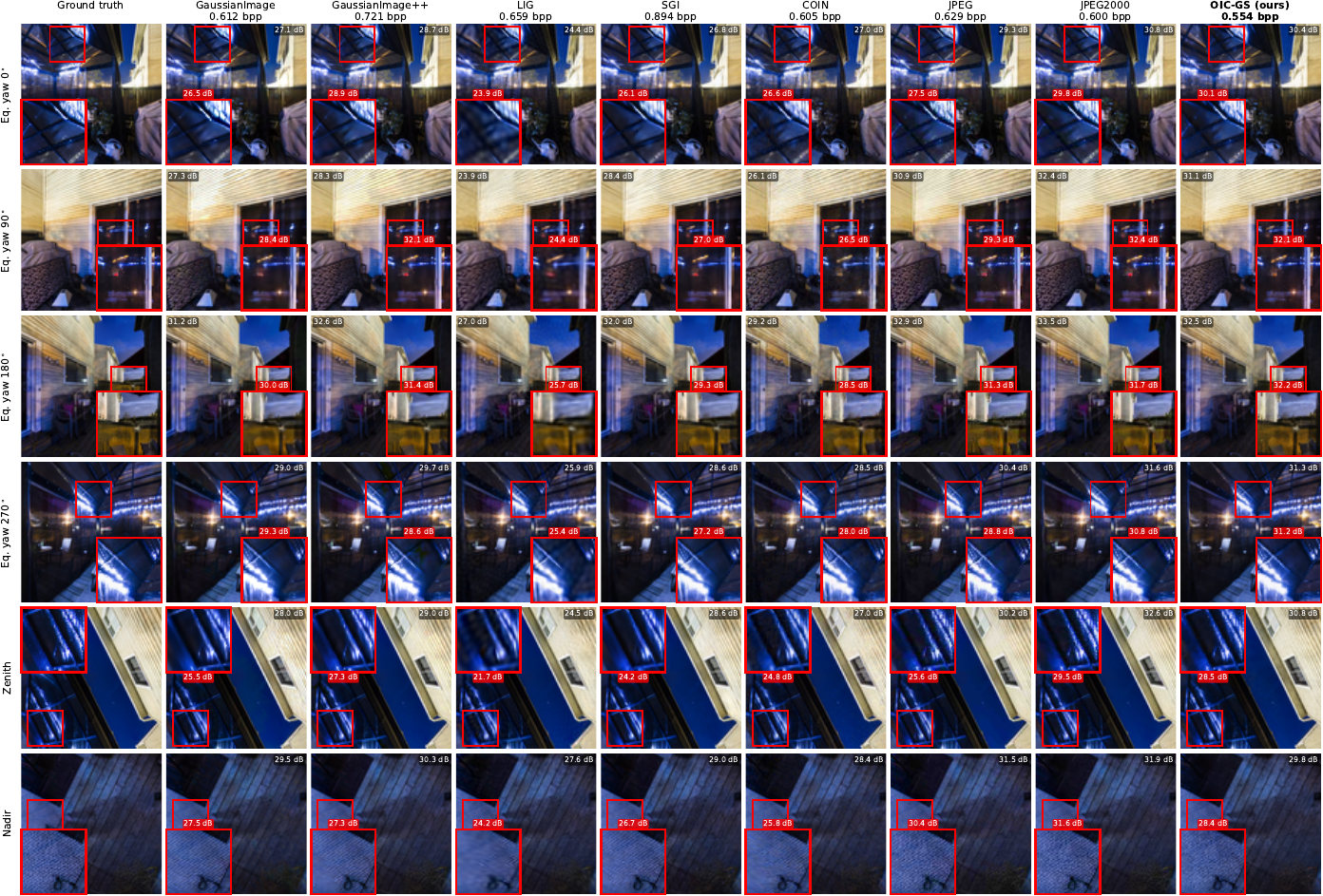}\\[3pt]
\includegraphics[width=0.92\textwidth]{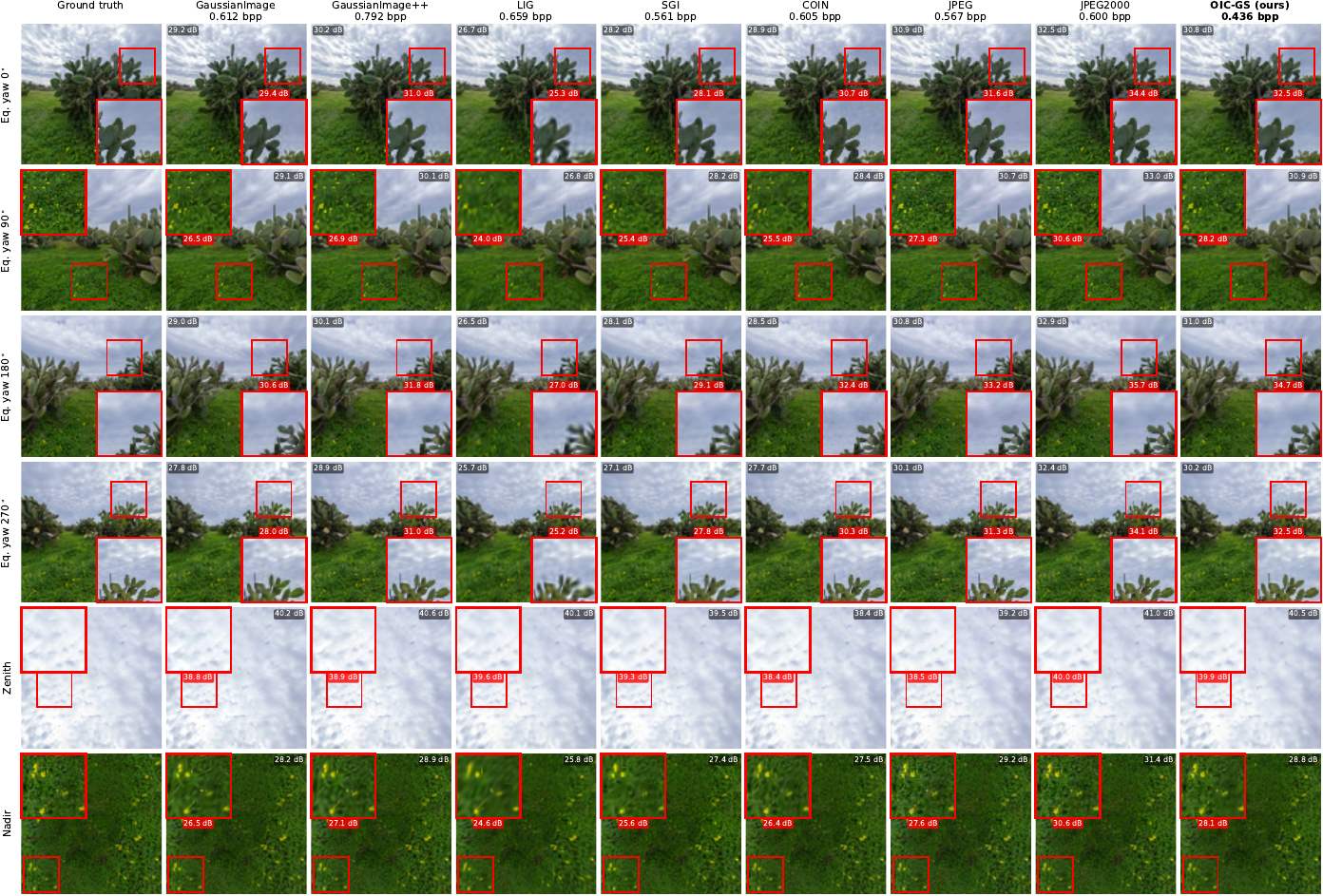}
\caption{Additional qualitative results (part 2 of 3): scenes 74 (top) and 46 (bottom) in all six viewing directions, under the protocol of Fig.~\ref{fig:qualgrid1}. Every baseline uses more bits than \method{} on both scenes ($0.554$ against $0.600$-$0.894$\,bpp on scene 74, and $0.436$ against $0.561$-$0.792$\,bpp on scene 46). Four of the five panels in Figs.~\ref{fig:qualgrid1}--\ref{fig:qualgrid3} where a GS codec exceeds \method{} in viewport PSNR are shown here, all for GaussianImage++ at a $30$-$82\%$ higher bitrate and by $0.1$-$0.6$\,dB: the yaw-$180^\circ$ and nadir views of scene 74 and the zenith and nadir of scene 46. The nadir patch of scene 74 is the only panel where no textured window favors \method{} over JPEG; at the zenith of scene 55, the best window ties.}
\label{fig:qualgrid2}
\end{figure}

\begin{figure}[t]
\centering
\includegraphics[width=0.92\textwidth]{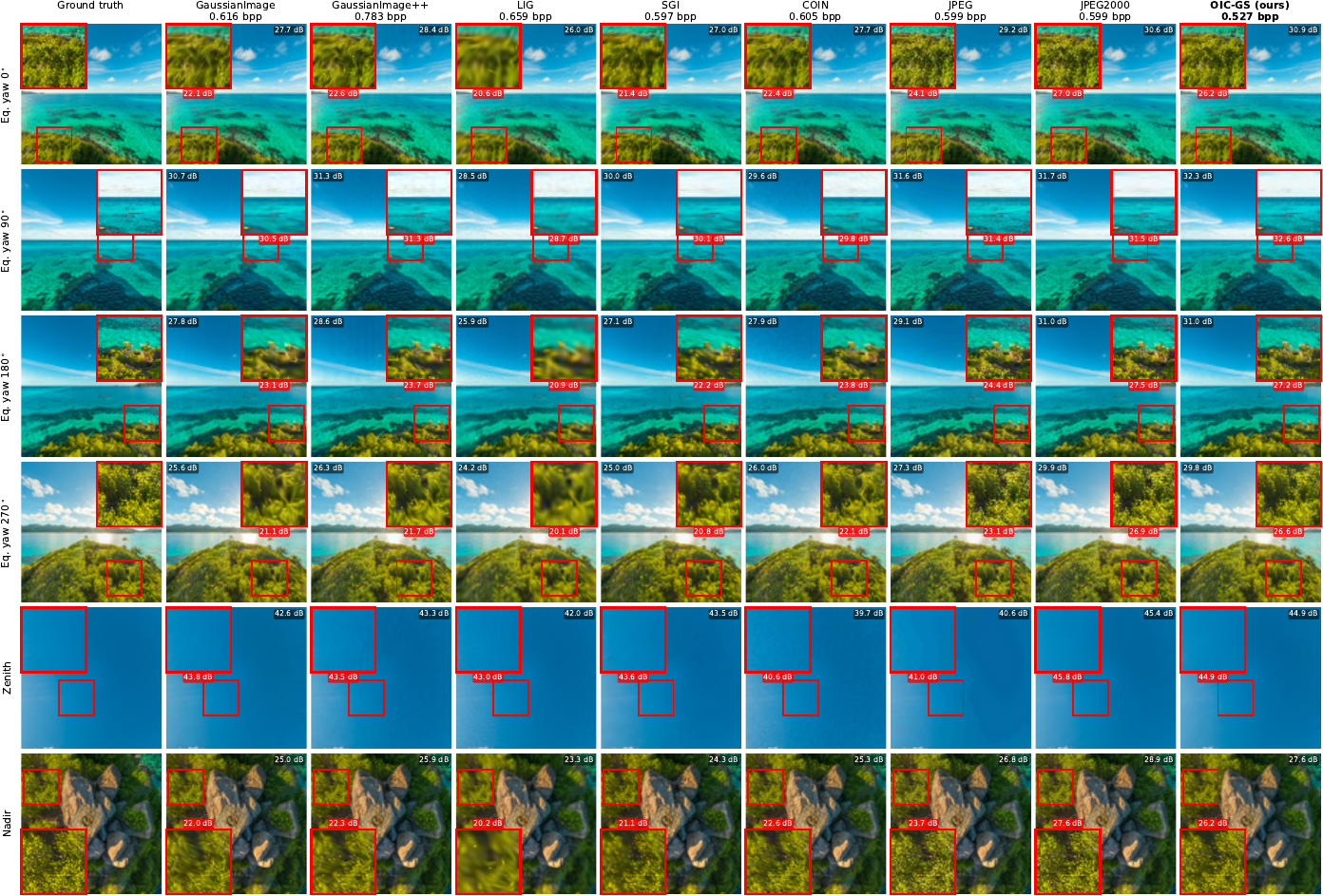}\\[3pt]
\includegraphics[width=0.92\textwidth]{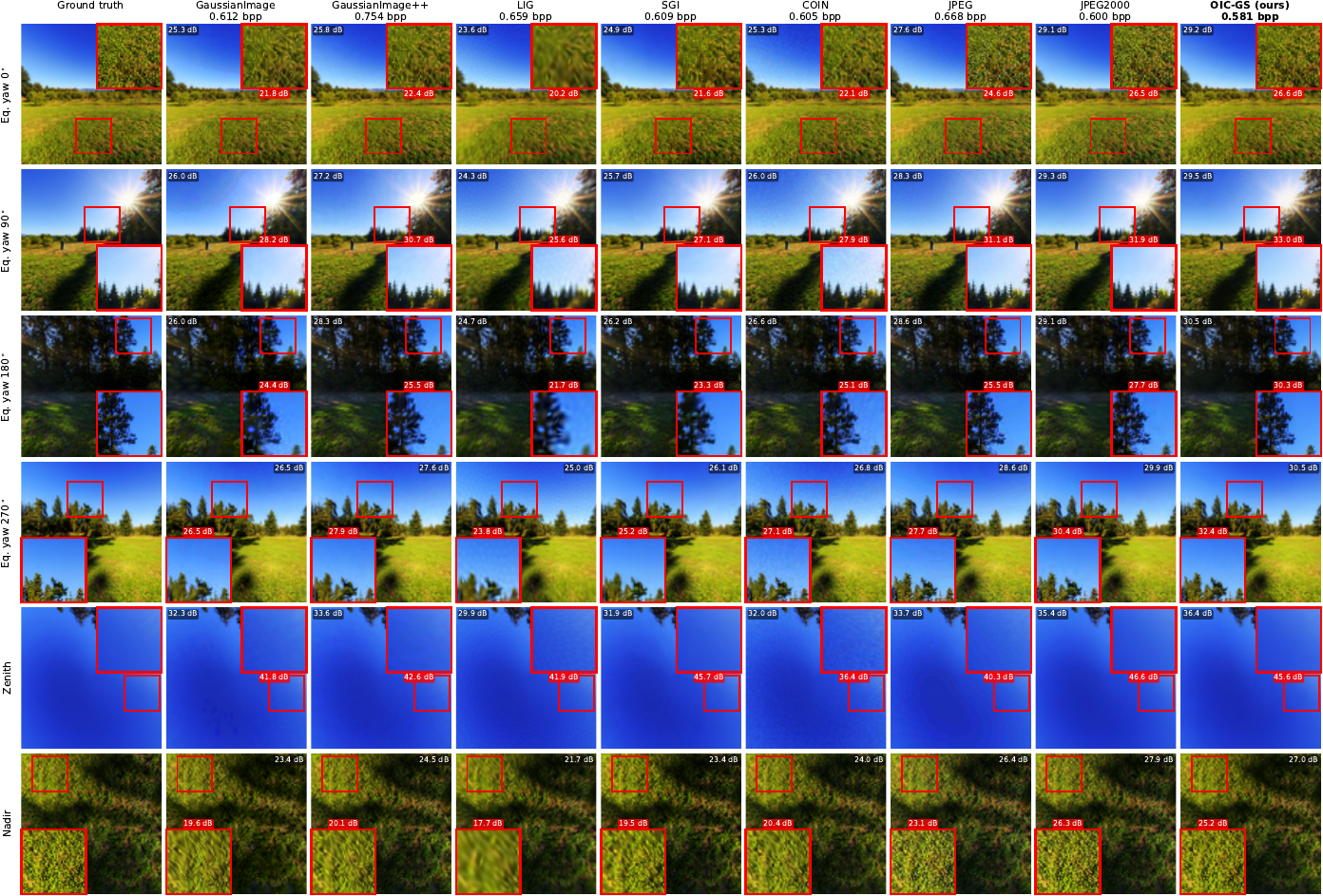}
\caption{Additional qualitative results (part 3 of 3): scenes 7 (top) and 32 (bottom) in all six viewing directions, under the protocol of Fig.~\ref{fig:qualgrid1}. Every baseline uses more bits than \method{} on both scenes ($0.527$ against $0.597$-$0.783$\,bpp on scene 7, and $0.581$ against $0.600$-$0.754$\,bpp on scene 32), and \method{} achieves the highest viewport PSNR among the GS codecs and COIN in all twelve panels.}
\label{fig:qualgrid3}
\end{figure}

Figs.~\ref{fig:qualgrid1}--\ref{fig:qualgrid3} extend Fig.~\ref{fig:qualitative} to five scenes and all six viewing directions, comparing against every method of Table~\ref{tab:bdrate} and against the JPEG anchor.
The column headers report the actual bitrate of each scene, and every baseline uses at least as many bits as \method{} on every displayed scene.
The scenes are the $10$th, $30$th, $50$th, $70$th and $90$th percentiles of the per-image WS-PSNR gain of \method{} over SGI ($+1.2$-$+3.0$\,dB), and were selected before inspecting the results.
The $90^\circ{\times}90^\circ$ viewports follow the projection protocol of App.~\ref{app:repro_eval}; labels and zoomed-in patches follow Fig.~\ref{fig:qualitative}.
A window in which \method{} outperforms JPEG exists in $28$ of the $30$ panels, and the best window ties at the zenith of scene 55; for the remaining panel, the nadir of scene 74, we show the most textured window.
Showing all six directions compares each panorama in full. Some regions, such as the sky at the zenith, are largely uniform and contain mainly low-frequency content, so the methods differ little there; in textured regions such as grass and buildings, \method{} recovers noticeably more high-frequency detail, as the zoomed-in patches show.

\end{document}